\documentclass[conference]{IEEEtran}

\usepackage{my_style}
\newacronym{UAV}{UAV}{Unmanned Aerial Vehicle}
\newacronym{ML}{ML}{Machine Learning}
\newacronym{TIR}{TIR}{Thermal Infrared}
\newacronym{DFSC}{DFSC}{Dual Flow Semantic Consistency}
\newacronym{mAP}{mAP}{Mean Average Precision}
\newacronym{SA}{SA}{State Accuracy}
\newacronym{mSA}{mSA}{Mean State Accuracy}
\newacronym{OV}{OV}{Out of View}
\newacronym{OC}{OC}{Occlusion}
\newacronym{FM}{FM}{Fast Motion}
\newacronym{SV}{SV}{Scale variation}
\newacronym{LI}{LI}{Low Illumination}
\newacronym{TC}{TC}{Thermal Cross-over}
\newacronym{IR}{IR}{Infra-Red}
\newacronym{LR}{LR}{Low Resolution}
\newacronym{CNN}{CNN}{Convolutional Neural Network}
\newacronym{CSM}{CSM}{Class-level Semantic Modulation}
\newacronym{ISM}{ISM}{Instance-level Semantic Modulation}
\newacronym{DL}{DL}{Deep Learning}
\newacronym{CF}{CF}{Correlation Filter}
\newacronym{FPS}{FPS}{Frame Per Second}
\newacronym{UTT}{UTTracker}{Unified Transformer-based Tracker}
\newacronym{MRLT}{MRLT}{Multi-Region Local Tracking}
\newacronym{GD}{GD}{Global Detection}
\newacronym{BC}{BC}{Background Correction}
\newacronym{DSOD}{DSOD}{Dynamic Small Object Detection}
\newacronym{RMCM}{RMCM}{Robust Motion Constraint Module}
\newacronym{FSRM}{FSRM}{Flexible Spatial Remapping Module}
\newacronym{ATUS}{ATUS}{Adaptive Template Update Strategy}
\newacronym{MCJT}{MCJT}{Multi-Consistency Joint Tracker}
\newacronym{AUC}{AUC}{area under the curve}
\makeglossaries

\begin{document}
\onecolumn
\setlength{\textfloatsep}{8pt}

\title{Performance Evaluation of Deep Learning-based Quadrotor UAV Detection and Tracking Methods\\
\thanks{All authors are with King Fahd University for Petroleum and Minerals, Dhahran, 31261, Saudi Arabia.\\
$^{\star}$ Equally contributed.}
}
\author{Mohssen E. Elshaar$^{\star}$, Zeyad M. Manaa$^{\star}$, Mohammed R. Elbalshy$^{\star}$, \\ Abdul Jabbar Siddiqui, and Ayman M. Abdallah
}
\maketitle

\begin{abstract}

\glspl{UAV} are becoming more popular in various sectors, offering many benefits, yet introducing significant challenges to privacy and safety. This paper investigates state-of-art solutions for detecting and tracking quadrotor \glspl{UAV} to address these concerns. Cutting-edge deep learning models, specifically the YOLOv5 and YOLOv8 series, are evaluated for their performance in identifying \glspl{UAV} accurately and quickly. Additionally, robust tracking systems, BoT-SORT and Byte Track, are integrated to ensure reliable monitoring even under challenging conditions. Our tests on the DUT dataset reveal that while YOLOv5 models generally outperform YOLOv8 in detection accuracy, the YOLOv8 models excel in recognizing less distinct objects, demonstrating their adaptability and advanced capabilities. Furthermore, BoT-SORT demonstrated superior performance over Byte Track, achieving higher IoU and lower center error in most cases, indicating more accurate and stable tracking.

\noindent Code: \href{https://github.com/zmanaa/UAV_detection_and_tracking}{https://github.com/zmanaa/UAV\_detection\_and\_tracking}\\
Tracking demo: \href{https://drive.google.com/file/d/1pe6HC5kQrgTbA2QrjvMN-yjaZyWeAvDT/view?usp=sharing}{https://drive.google.com/file}\\
\end{abstract}

\begin{figure}[H]
    \centering
    \includegraphics[width=\linewidth]{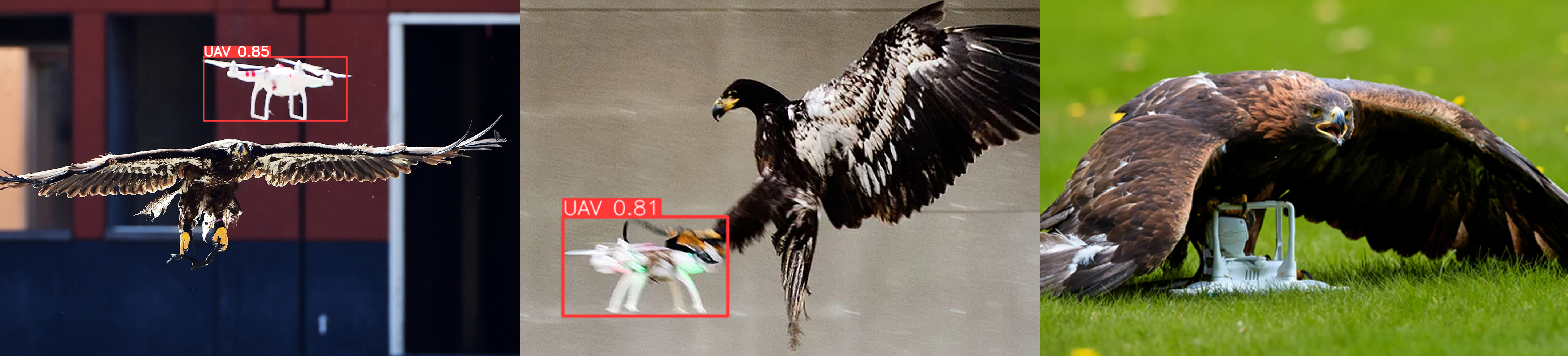}
    \caption{Hawks attacking UAVs, illustrating a potential natural anti-drone defense mechanism.}
    \label{fig:Cover}
\end{figure}

\section{Introduction}
\label{sec:introduction}

\glspl{UAV} have garnered a lot of interest recently due to their accessibility and usefulness \cite{lort2017initial}. UAVs were originally created for military applications, but they are now used in a variety of industries, including transportation \cite{xu2017car}, environmental monitoring \cite{sharma2016uav, asadzadeh2022uav}, and logistics \cite{vskrinjar2019application}. \glspl{UAV} have advantages, but they also have disadvantages, most notably when it comes to privacy, personal safety, and public safety. The growing use of \glspl{UAV} has given rise to a number of problems, such as threats to privacy, individual safety, and public safety. Thus, it is essential to build efficient systems to identify and monitor unintentional or undesired \gls{UAV} intrusions. There isn't a completely reliable anti-\gls{UAV} tracking and detection technology available just yet. The majority of detection and early warning systems in use today utilise radar, radio frequency (RF), and acoustic sensors \cite{hoffmann2016micro, abunada2020design, chang2018surveillance}. These systems frequently have flaws, such as high cost and noise susceptibility, that lead to inaccurate findings. As such, these algorithms are limited to use in public spaces such as airports. Therefore, it is imperative to detect and monitor any \glspl{UAV} that may be unintentionally or illegally invading. Nevertheless, anti-\gls{UAV} detection remains a challenging problem, with no consistently reliable method to date.

Currently in use, most detection and early warning systems rely on acoustic, radio frequency (RF), and radar sensors \cite{hoffmann2016micro, abunada2020design, chang2018surveillance}. Frequently, these systems have shortcomings that result in erroneous results, such as excessive cost and susceptibility to noise. This means that these algorithms can only be used in public places like airports. The detection and surveillance of inadvertent or illicit \glspl{UAV} encroachment are therefore critical. Still, there is no widely accepted, trustworthy mechanism for anti-\gls{UAV} detection, making the problem difficult.

Deep learning techniques have advanced quickly in the last few years in a number of computer vision fields \cite{gao2018large, gao2020cross, gao2020hierarchical, gao2021constructing, ren2015faster, bertinetto2016fully}, especially in object tracking and identification. These techniques are mature enough to provide a high-performing anti-UAV tracking system. There are presently a number of standard tracking models including SiamFC \cite{bertinetto2016fully} and DiMP \cite{bhat2019learning}, as well as generic object identification models like Faster-RCNN \cite{ren2015faster} and SSD \cite{liu2016ssd}. Nevertheless, straight application of these generic methods to UAV tracking and identification yields poor results. The main focus of anti-UAV detection is still small target detection against complicated backgrounds, despite the detection algorithms' progressive development and commercialization. \glspl{UAV} produce a lot of noise and interference since they frequently blend in with the complex surroundings. Furthermore, occlusion presents difficulties for the tracking process. Numerous strategies have been proposed to overcome these problems and produce positive outcomes. For example, YOLOv3 has been improved \cite{hu2019object} and low-rank and sparse matrix decomposition has been used for classification \cite{wang2019flying}.

Significant research and development in anti-UAV systems has occurred recently due to growing concerns about the safety of unmanned aerial vehicles (UAVs), especially in the context of national security. Numerous nations have made significant investments in sophisticated anti-UAV systems, mostly found in military installations, that do not rely on deep learning. These systems are being actively improved by universities and research centers.

To identify, locate, and protect against drones, \cite{shi2018anti} proposed the ADS-ZJU system that combines a number of surveillance technologies. To gather RF signals, video images, and auditory signals, it uses three sensors. A central unit processes these inputs and extracts information for localization and detection. ADS-ZJU uses the short-time Fourier transform to analyze the spectrum of acoustic signals, describes image features using histograms of oriented gradients, and separates Wi-Fi signals from UAV RF signals. Support vector machines (SVM) are used for parallel detection of RF, video, and audio signals. Based on video images, hybrid measurements are utilized to estimate the location of the UAV, such as received signal strength (RSS) and direction of arrival (DOA). The system can handle radio frequency interference and achieve excellent accuracy by merging different surveillance methods. However, because of its expensive cost, the system is more suited for military usage than civilian use because its dispersed units cover a broad region.

The work of \cite{sheu2019development} proposed the Dynamic Coordinate Tracing method which suggests a dual-axis rotating tracking mechanism that measures the UAV's flying altitude by using a tracing device fitted with full-color or thermal imaging cameras and sensing modules. The device dynamically determines the coordinates of the UAV in spherical coordinates, taking latitude and longitude into account. The system can use either thermal imaging or full-color cameras to adjust to varying weather conditions. This tracking device provides anti-UAV systems with a useful and affordable option. For it to function properly, though, top-notch hardware facilities are still needed.

\subsection{Contribution}
To this end, we make the following contributions: \begin{itemize}
 
    \item \textbf{Benchmarking State-of-the-Art Models:} We evaluate and compare four versions of YOLOv5 \cite{ultralytics2024yolov5} and four versions of YOLOv8 \cite{ultralytics2024yolov8} for UAV detection and tracking tasks using RGB images.
    \item \textbf{Comprehensive Evaluation Framework:} We establish a benchmark framework that systematically assesses the performance of different object detection and tracking models in various scenarios, including challenging environments with complex backgrounds and occlusions.
    \item \textbf{Publicly Accessible Resources:} We provide trained model weights and demonstration code for each detection model, as well as a tracker, available through our GitHub repository: \href{https://github.com/zmanaa/UAV_detection_and_tracking}{https://github.com/zmanaa/UAV\_detection\_and\_tracking}.
    \item \textbf{Performance Analysis:} We analyze the strengths and limitations of each model, offering insights into their suitability for real-time anti-\gls{UAV} detection and tracking applications.
    \item \textbf{Novel Experimental Insights:} We present unique insights derived from our experiments that, to the best of our knowledge, have not been previously reported in the computer vision literature.
    \item \textbf{Recommendations for Future Research:} Based on our findings, we propose directions for enhancing UAV detection and tracking systems, including potential model improvements and integration strategies.
\end{itemize}
Additionally, we organise the paper as follows:
\begin{inparaenum}[i)]
    \item {Section \ref{sec:related-work}:} Provides an overview of relevant related works and datasets.
    \item {Section \ref{sec:methodology}:} Details the methodology adopted for performance evaluation studies.
    \item {Section \ref{sec:results-discussions}:} Presents the results and discussions.
    \item {Section \ref{sec:conclusion}:} Concludes the paper and outlines directions for future work.
\end{inparaenum}





\section{Related Work}
\label{sec:related-work}

\color{black}
In this section, a brief review of recent related works on the problem of UAV/drone detection and tracking is provided. These works could be categorized into: (i) RGB Images-based Tracking, (ii) RGB and Depth Images-based Tracking, (iii) RGB and Thermal Images-based Tracking, (iv) Thermal Images-based Tracking, (v) Object Detection and Tracking, (vi) Hybrid Anti-UAV Systems. Moreover, we review some datasets relevant to the issue of UAV detection and tracking.
\color{black}

\subsection{RGB-based Tracking}
Over the past ten years, tracking techniques based on Red, Green and Blue (RGB) color information have significantly improved. \color{black}In RGB-based tracking, the input typically consists of visual data, such as images or video frames, where each pixel is represented by its RGB color values. \color{black} Several methods have produced good results in short-term tracking, including correlation filtering-based trackers \cite{bolme2010visual, danelljan2015learning, henriques2014high}. 
Additionally, by changing the tracking problem into a similarity-matching problem, Siamese/transformer-based trackers \cite{cen2018fully, li2018high, chen2021transformer, zhou2022global} have become more and more popular. While these trackers—SiamFC, SiamRPN, and SiamFC++—have demonstrated good accuracy, they have difficulty meeting real-time needs. 
Target tracking has advanced thanks to the emergence of benchmark datasets like OTB, LaSOT, UAV123, and others \cite{Wu2013OnlineObjectTracking, fan2019lasot, mueller2016benchmark}.

\subsection{RGB and Depth-based Tracking}
RGB and Depth (RGBD) tracking techniques have drawn interest 
by complementing RGB data by using low-cost depth cameras to capture precise spatial information from depth photos. This method boosts tracking performance and solves problems like occlusion in an efficient manner. Techniques such as CA3DMS \cite{liu2018context} use 3-D mean-shift approaches to address occlusion issues, while OTR \cite{kart2019object} builds a spatial reliability map based on color and depth information to enable effective 3-D target model reconstruction. There are two types of RGBD tracking approaches: early fusion and late fusion schemes. While late fusion analyzes and decides individually for each modality, early fusion integrates features from both RGB and depth modalities. To assess RGBD tracking techniques, benchmark datasets such as PTB, STC, CDTB, and DepthTrack \cite{song2013tracking, yan2021depthtrack, xiaorobust, wang2020robust} have been produced. Although deep learning models do exceptionally well in tracking, their applicability in real-world scenarios may be limited by their inability to cope with temporal and spatial disruptions.

\subsection{RGB and Thermal Images-based Tracking}
The combination of thermal infrared (TIR) and RGB modalities, known as RGBT tracking, has drawn interest. The three primary types of RGBT tracking are deep learning-based approaches, correlation filter-based approaches, and sparse representation-based methods. To accomplish robust RGBT tracking, early methods relied on sparse representation and included data fusion, modal weight computation, and Bayesian filtering \cite{wu2011multiple, li2016learning, li2017weighted}. Moreover, correlation filtering methods that combine RGB and TIR modalities and make use of global suggestion and local sampling techniques have been investigated \cite{xu2021drf, jun2022rgb}. Using strong feature representations, deep learning techniques like mfDiMP and CIRNet have become popular \cite{zhang2019multi, xia2022cirnet}. The development of RGBT tracking is hampered by a lack of training data, and the benchmark datasets that are now available include GTOT, RGBT210, and RGB234 \cite{li2016learning, li2017weighted, li2019rgb}.

\subsection{TIR based tracking}
Traditional TIR tracking methods rely on handcrafted features such as HOG \cite{dalal2005histograms,camplani2015real} and gray-scale information to track the target. Variants of these methods have been developed to address various challenges. These include noise reduction techniques, algorithms to handle changes in target scale, and approaches to mitigate the effects of brightness and contrast changes. To overcome the limitations of lacking color information and vague edge structure, researchers have explored the use of appearance models based on intensity histograms \cite{venkataraman2012adaptive} and temperature-based mean displacement \cite{yun2019tir} algorithms. Other advancements include the development of algorithms based on distributed field representation\cite{berg2016channel}, as well as temperature-based mean-shift \cite{yun2019robust} and mask-based trackers \cite{li2019mask}. However, these traditional methods often exhibit poorer tracking performance compared to other framework-based trackers due to their reliance on simple feature extraction and limited consideration of intensity characteristics.
 Correlation Filters-Based TIR tracking methods offer a more robust framework for TIR tracking. They utilize the initial frame and expected label of the target to train a filter model. By convolving features extracted from the search area with the trained correlation filter, a response map is generated. The target's location is determined by locating the maximum point in the response map \cite{yuan2023thermal} \citep{henriques2014high, yuan2020robust, luo2019thermal, kiani2017learning, yuan2020trbacf}. Scale evaluation can be performed using a pyramid with multiple scale factors, and model update techniques adjust parameters to accommodate target changes \cite{yuan2023thermal}. Researchers have improved tracking performance in this category by incorporating weighted multiple features \cite{he2015infrared} and utilizing convolutional features\cite{danelljan2014accurate}, which provide richer information. Advanced models such as ECO-LS and LMSCO \cite{gao2018large,liu2020learning} have been introduced to address challenges such as deformation, occlusion, and accurate scaling. These methods have shown promising results and aim to enhance the accuracy, robustness, and efficiency of TIR target tracking systems. 
 Various data sets have been found in the literature to be used in TIR tracking task. Among these, the datasets from OSU \cite{davis2007background}, LITIV \cite{torabi2012iterative}, ASL-TID \cite{portmann2014people}, and BUTIV \cite{smeulders2013visual} are out of date and impractical for certain applications, like short-term tracking of a single target. The VOT-TIR15 \cite{felsberg2015thermal}, VOT-TIR16 \cite{lebeda2016thermal}, VOT-TIR17 \cite{kristan2015visual}, PTB-TIR \cite{liu2019ptb}, and LSOTB-TIR \cite{liu2020lsotb} datasets, on the other hand, are widely recognized and frequently used to assess the effectiveness of TIR trackers. These datasets are useful reference points for evaluating the precision and efficacy of TIR tracking techniques. 

\subsection{UAV tracking and detection}

Corresponding algorithms have also been developed. Particular difficulties arise while detecting and tracking UAVs from an aerial perspective, including densely populated areas, tiny objects, and intricate backdrops. Exchange Object Context Sampling (EOCS) is one technique used to overcome these issues by taking into account object relationships and contextual information \cite{4yu2020unmanned, 5yu2018online}. To manage quick camera motion, optimization of camera motion models based on backdrop feature points has been suggested \cite{6li2017visual}.
Furthermore, a lightweight Transformer layer has been incorporated into pyramid networks to produce a real-time CPU-based tracker, taking into account the restricted computational capabilities on UAVs \cite{7xing2021siamese}. Due to these algorithms' strong performance on current UAV tracking benchmarks, airborne object tracking is becoming more widely available for commercial use. The significance of anti-UAV tracking is further highlighted by the growing popularity of UAV tracking \cite{4yu2020unmanned, 6li2017visual, 7xing2021siamese}.

\subsection{Datasets}

{\color{red}

\color{black}
UAV perspective on object recognition and tracking is currently gaining more attention. UAVs are appropriate for airborne object monitoring because they provide more control and flexibility than cameras mounted on moving vehicles. To address these challenges, a number of UAV datasets have been generated, notably UAV123 for tracking and DroneSURF and CARPK \cite{mueller2016benchmark, 2kalra2019dronesurf, 3hsieh2017drone} for detection \cite{4yu2020unmanned, 7xing2021siamese}.
\color{black}
Deep learning-based object tracking algorithms are now used for UAV tracking, supplementing existing detection techniques, thanks to advances in computer vision. The availability of datasets is essential for training models and ensuring resilience. Several noteworthy UAV datasets have been created, including:
\subsubsection{MAV-VID}
 This Kaggle collection of 64 movies (40,323 pictures) \cite{rodriguez2020adaptive} is devoted only to the detection of UAVs. The UAVs are modest, averaging 0.66\% of the total image, primarily horizontally scattered, and relatively concentrated in particular areas. Our dataset, on the other hand, has a dispersed distribution of UAVs with more consistent vertical and horizontal distributions, giving the trained models more resilience.
 \subsubsection{Drone-birds dataset \cite{coluccia2019drone}}
 Presented at the 16th IEEE International Conference on AVSS, this dataset features birds and unmanned aerial vehicles (UAVs) as objects of interest. Due to their similar sizes, colors, and shapes, it can be difficult to distinguish between drones and birds. This version of the dataset includes both land and sea scenes that were shot using various cameras. The average size of the observed UAVs in this collection is 34x23 pixels, or 0.1\% of the total image size. There are 77 videos with over 10,000 photos accessible. The dataset holds importance in enhancing algorithms to address false positives and perhaps implementing them in different fields. This dataset's scenes mostly show beaches with a broad field of view, but our collection is more appropriate for civilian use because it concentrates on urban settings.
 \subsubsection{AntiUAV \cite{Jiang2021AntiUAV}}
 This dataset includes 318 fully labeled films and provides labeled dual-mode information for both visible and infrared light. With 186,494 images altogether, it consists of three sets: 91 videos for testing, 160 films for training, and 160 videos for validation. This dataset's UAVs are divided into seven attributes that address different unique situations that arise during UAV detection missions. The two modalities—day and night—are given distinct roles in the videos that are captured in the dataset. The anti-UAV dataset shows less volatility than previous datasets, including ours, and offers wide-ranging motion, albeit largely in the central region. While the nighttime scenarios in this dataset are the main focus, our dataset strives to improve model robustness by adding several factors such UAV kinds, scene information, lighting conditions.

 \subsubsection{DUT dataset}
 In order to promote progress in UAV tracking and detection, the DUT Anti-UAV  dataset was developed by \cite{zhao2022vision}. There are two subgroups in this dataset: tracking and detection. The tracking subset consists of 20 sequences with various UAV targets, while the detection subset is separated into training, testing, and verification sets. A random sample from the data set can be seen in Fig. \ref{fig:dut_sample}.  More on this data set will be discussed in section \ref{sec:data_struc}. 
 
 Table \ref{Comparison against MAV, Drone-Bird, Anti-UAV, and DUT datasets} compares between MAV, Drone-Bird, Anti-UAV, and DUT datasets in terms of number of videos, number of images, target size to total image, UAV types, in addition to light modes, light conditions, weather conditions, backgpund and image resolution settings. The table indicates that DUT datasets has more complex and diverse backgrounds in the images and also uses diverse light and weather conditions. These variations in the dataset enrich its diversity and help in solving the problem of model overfitting. Moreover, The DUT dataset's complex background and noticeable changes in outdoor lighting are essential for developing a robust ,reliable, and effective UAV detection model. DUT also consider various settings of image resolution which ease the adaptation to images with different sizes, and also helps in overfitting avoidance.

\begin{table}[]
\caption{Comparison between MAV, Drone-Bird, Anti-UAV, and DUT datasets}
\label{Comparison against MAV, Drone-Bird, Anti-UAV, and DUT datasets}
\resizebox{\textwidth}{!}{%
\begin{tabular}{@{}p{1.4in}p{1in}p{1in}p{1in}p{2in}@{}}
\toprule
                           & MAV \cite{rodriguez2020adaptive} & Drone-Bird \cite{coluccia2019drone} & Anti-UAV \cite{Jiang2021AntiUAV}                          & \textbf{DUT}  \cite{zhao2022vision} \checkmark
                            \\ \midrule
No. videos                 & 64                  & 77                                & 318 video                          & 20                                           \\
No. images                 & 40,323              & 10,000                            & 186,494                            & 24,804\\
Target size to total image & 0.66\%              & 0.1\%                             & 0.4 to 0.5\%                       & object area ratio range from $1.9e^{-6}$ to 0.7 \\
UAV types                  & NA                  & NA                                & NA                                 & more than 354 types                          \\
light conditions           & NA                  & NA                                & day and night                      & day, night, dawn and dusk                    \\
light modes                & NA                  & NA                                & visible and infrared               & NA                                           \\
Weather conditions         & NA                  & NA                                & NA                                 & different weather (sunny, cloudy, and snowy day)                 \\
Background                 & NA                  & Sea side with a wide visual field & Diverse (buildings, clouds, trees, etc)                       & usually complicated ( the sky, dark clouds,
jungles, high-rise buildings, residential buildings, farmland,
and playgrounds)  \\
Image resolution           & NA                  & NA                                & NA                                 & Various settings of image resolution         \\ \bottomrule
\end{tabular}%
}
\end{table}

 For the aforementioned reasons and limitations of other datasets, we choose DUT dataset for our training process. The dataset is open access from 2022 and the authors of ref. \cite{zhao2022vision} have conducted a comprehensive study for different detection and tracking architectures. \color{black} Yet, the performance of different variations of YOLOv5 and YOLOv8 models was never tackled, as well as the tracking models provided by YOLOv8 has not been previously explored. 
In this work, we aim to provide an in-depth analysis of the performance of state-of-the-art object detectors and trackers using DUT dataset. While we do not introduce a new detection or tracking method, we conduct unique experiments that compare the YOLOv5 and YOLOv8 models, in addition to BoT-SORT and Byte Track tracking models provided by YOLOv8. These experiments, designed for previously untested scenarios, offer valuable insights into the strengths and limitations of these models.

\color{black}

 \begin{figure}[H]
     \centering
     \includegraphics[width = 0.5 \textwidth]{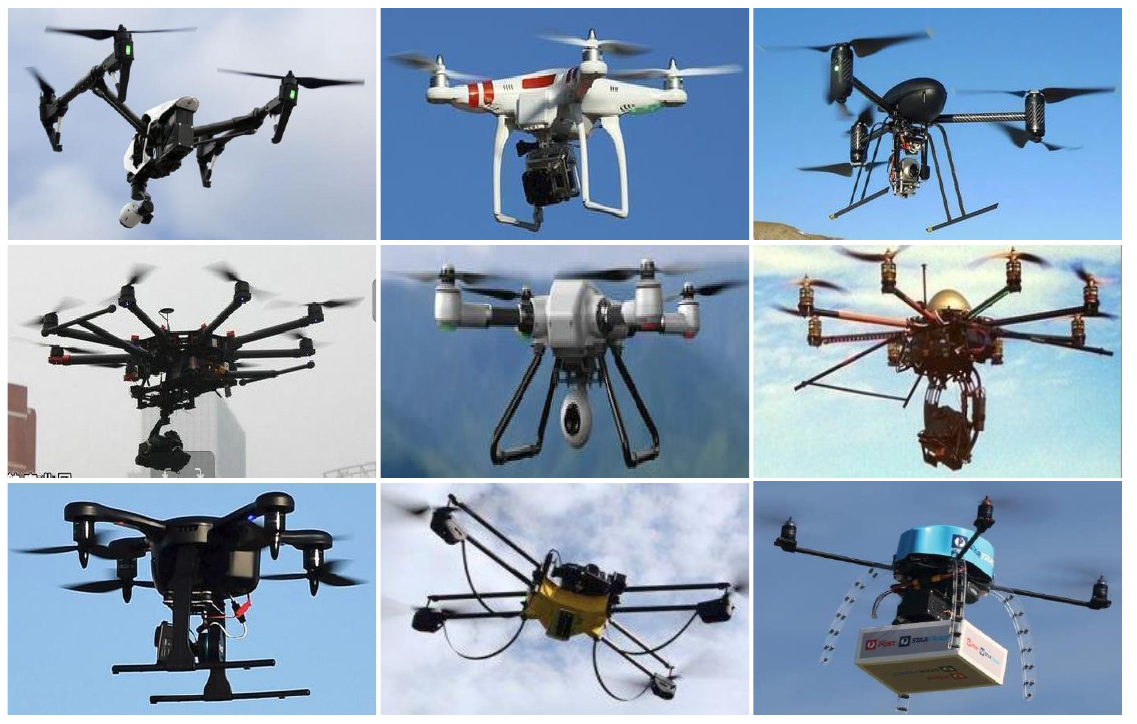}
     \caption{Samples from the DUT Anti-UAV dataset \cite{zhao2022vision}.}
     \label{fig:dut_sample}
 \end{figure}

\section{Experimental Methodology and Setup}
\label{sec:methodology}

In this section, we describe the methodology undertaken for the comprehensive performance evaluations of state-of-the-art deep learning models for the task of UAV detection and tracking.
First, we describe the DUT dataset used. Second, we describe the training and validation process. Third, the tracking models parameters are presented. 


\color{black}

\subsection{Data Structure} \label{sec:data_struc}
The DUT dataset \cite{zhao2022vision} is composed of independent detection and tracking datasets. The detection Dataset contains ten thousand images. 5200 are denoted for training, 2600 for validation and 2200 for testing. Each image file is accompanied by an \emph{.xml} file that includes tree structured data of the size of the image and its persisting objects Fig \ref{fig:DUT_Tree}. Information about the objects labels and bounding box (bbox) extremes ($x_{\text{min}}$, $y_{\text{min}}$, $x_{\text{max}}$ and $y_{\text{max}}$) could be extracted and transformed into suitable formats depending on the detection model. Statistics about the dataset can be found in Figs. \ref{fig:aspect_ratio}, \ref{fig:position}, and \ref{fig:area_ratio}.

\begin{figure}[]
    \centering
    \begin{subfigure}{0.3\textwidth}
        \centering
        \includegraphics[width = \textwidth]{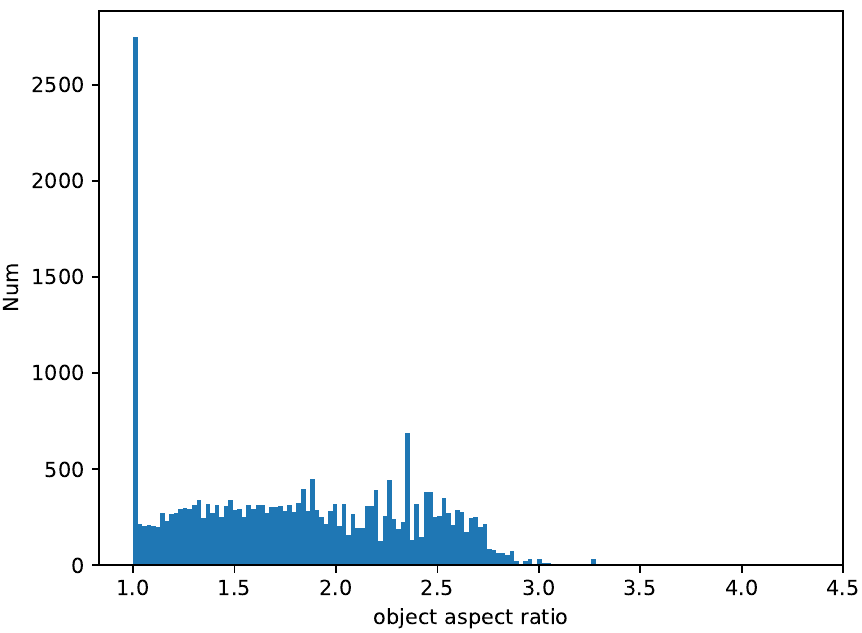}
        \caption{Train}
        \label{fig:yolov5s1}
    \end{subfigure}
    \begin{subfigure}{0.3\textwidth}
        \centering
        \includegraphics[width = \textwidth]{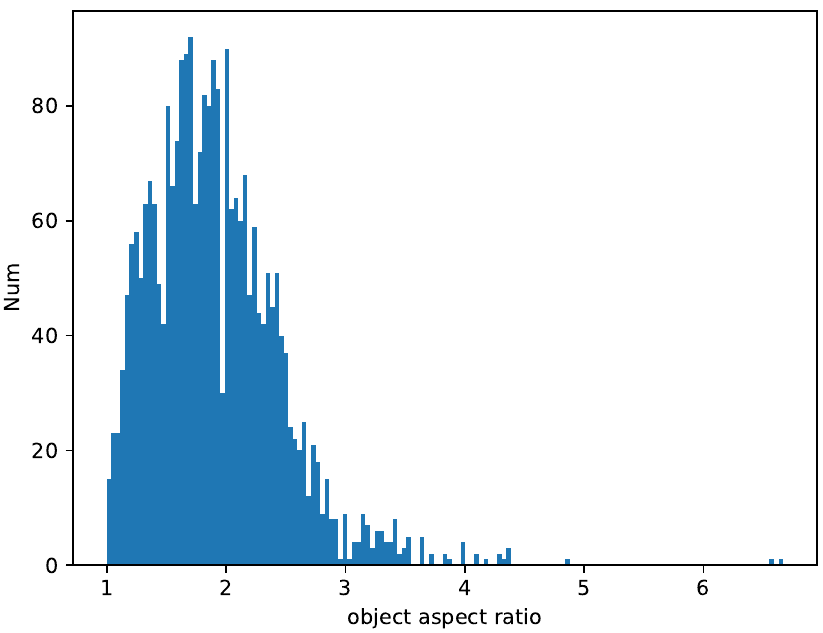}
        \caption{Validation}
        \label{fig:yolov5x1}
    \end{subfigure}
    \begin{subfigure}{0.3\textwidth}
        \centering
        \includegraphics[width = \textwidth]{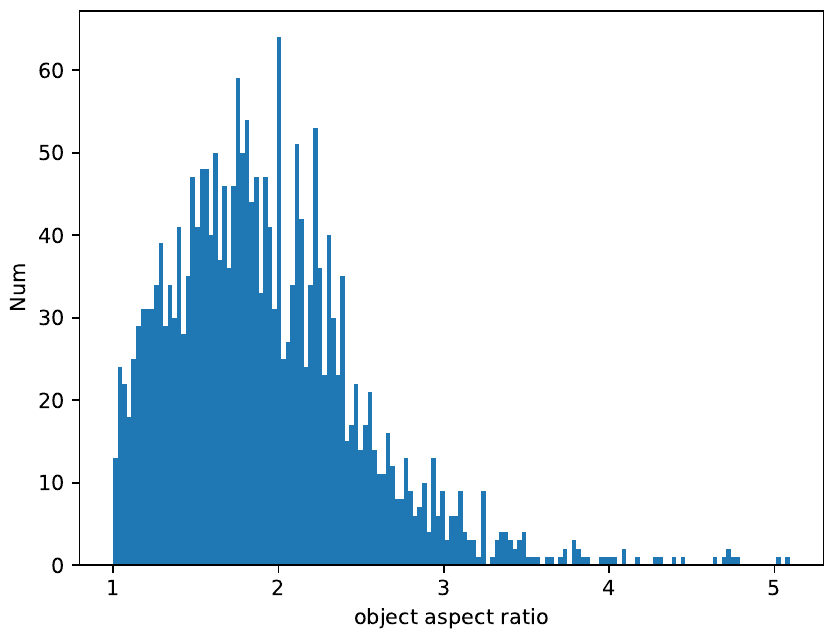}
        \caption{Test}
        \label{fig:yolov8s1}
    \end{subfigure}
    \caption{Aspect ratio statistics for the used images within the dataset.}
    \label{fig:aspect_ratio}
\end{figure}

\begin{figure}[]
    \centering
    \begin{subfigure}{0.3\textwidth}
        \centering
        \includegraphics[width = 0.8\textwidth]{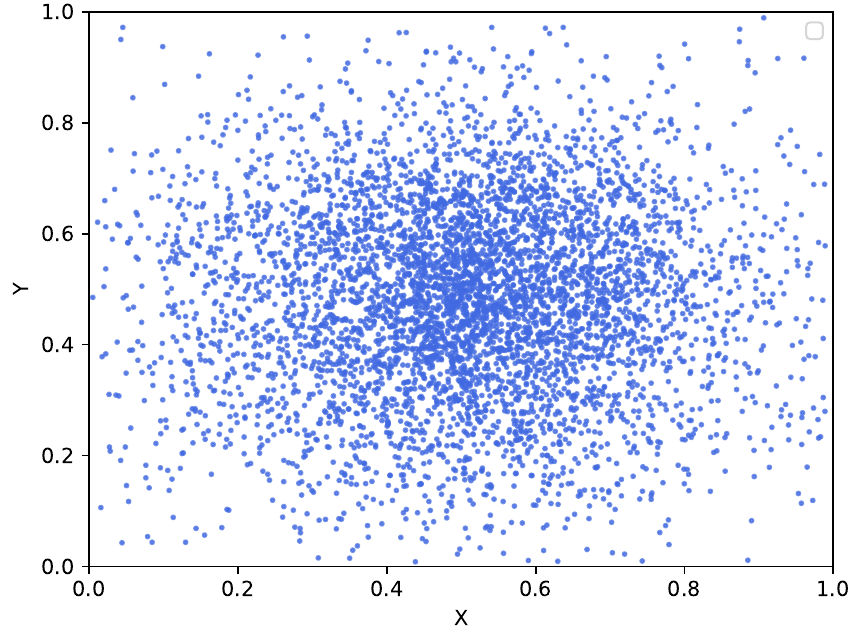}
        \caption{Train}
        \label{fig:yolov5s1}
    \end{subfigure}
    \begin{subfigure}{0.3\textwidth}
        \centering
        \includegraphics[width = 0.8\textwidth]{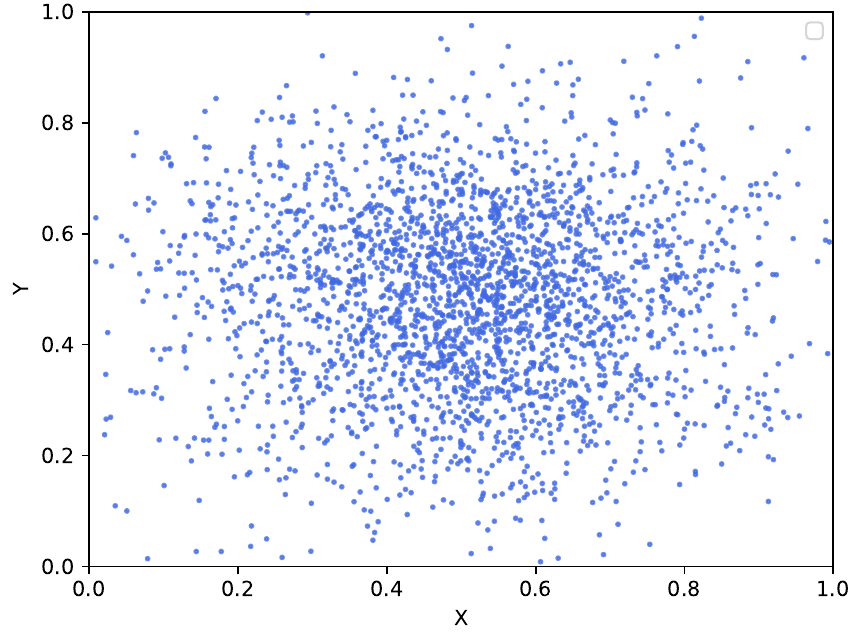}
        \caption{Validation}
        \label{fig:yolov5x1}
    \end{subfigure}
    \begin{subfigure}{0.3\textwidth}
        \centering
        \includegraphics[width = 0.8\textwidth]{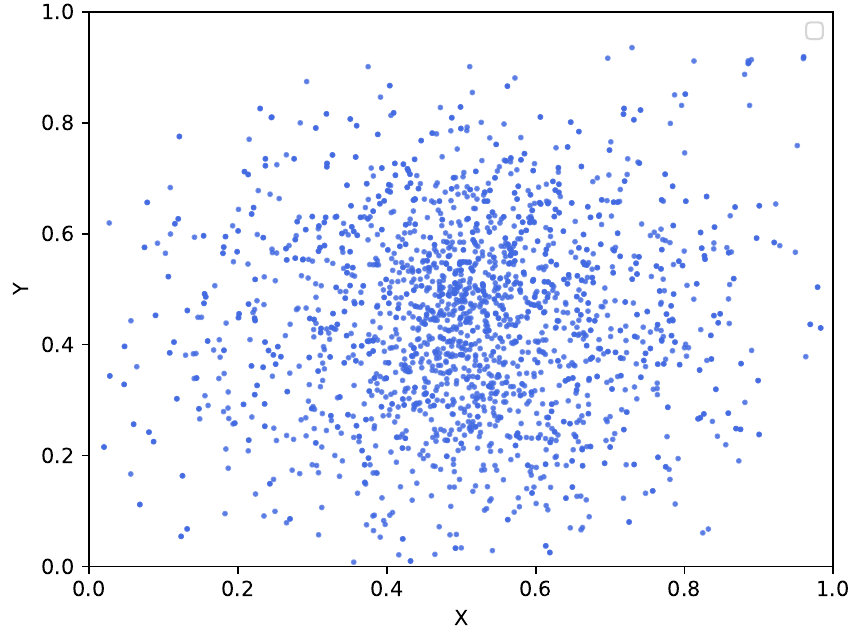}
        \caption{Test}
        \label{fig:yolov8s1}
    \end{subfigure}
    \caption{Position distribution of the object(s) within the used images in the dataset.}
    \label{fig:position}
\end{figure}

\begin{figure}[H]
    \centering
    \begin{subfigure}{0.3\textwidth}
        \centering
        \includegraphics[width = \textwidth]{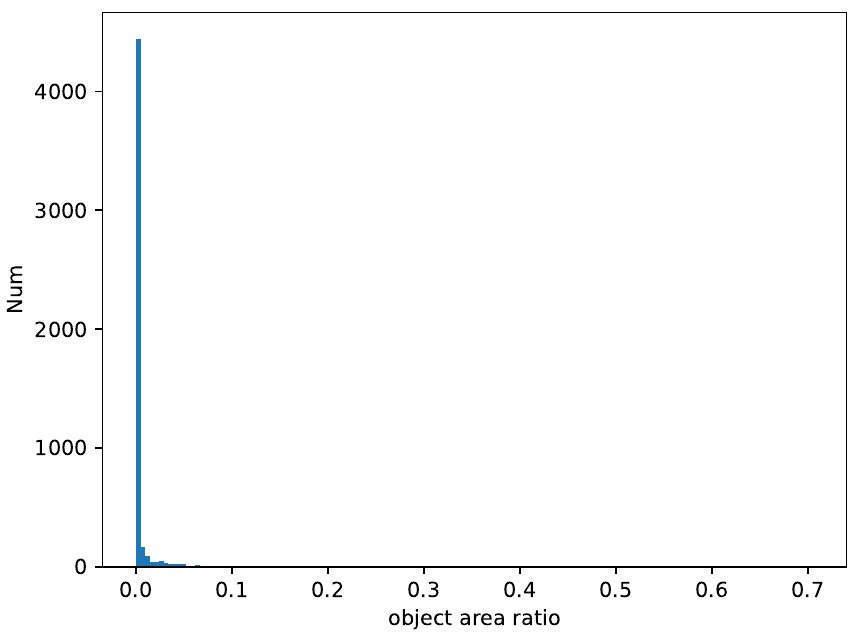}
        \caption{Train}
        \label{fig:yolov5s1}
    \end{subfigure}
    \begin{subfigure}{0.3\textwidth}
        \centering
        \includegraphics[width = \textwidth]{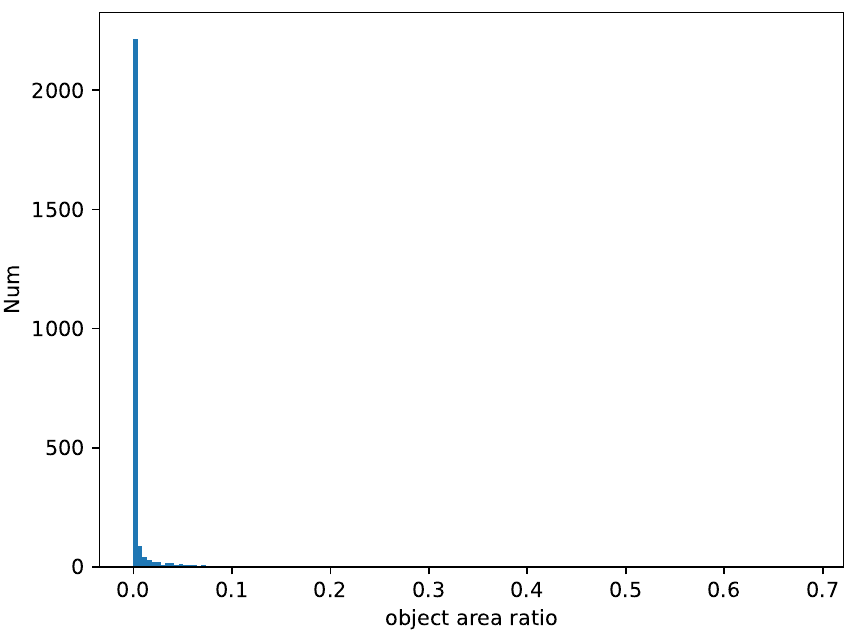}
        \caption{Validation}
        \label{fig:yolov5x1}
    \end{subfigure}
    \begin{subfigure}{0.3\textwidth}
        \centering
        \includegraphics[width = \textwidth]{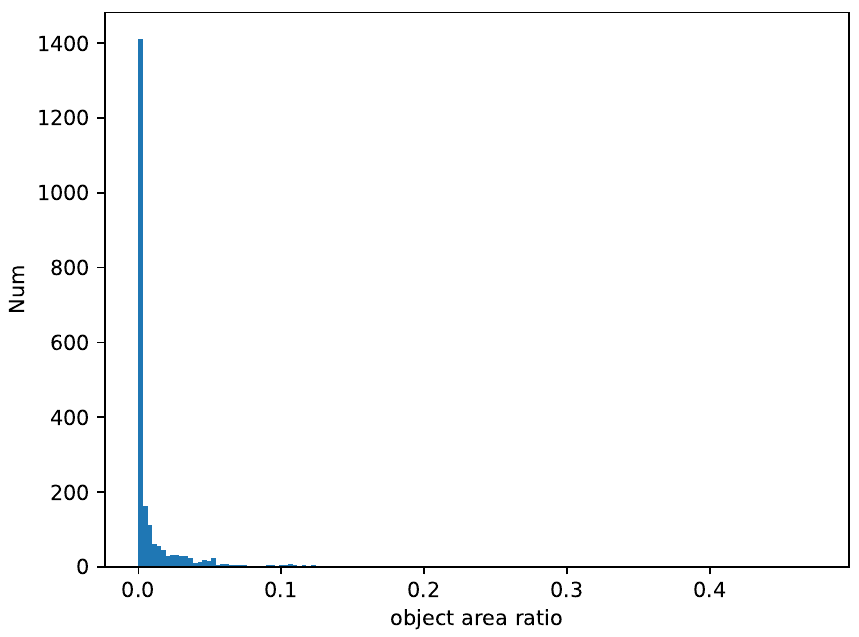}
        \caption{Test}
        \label{fig:yolov8s1}
    \end{subfigure}
    \caption{Area ratio between the object(s) and the image size in the dataset.}
    \label{fig:area_ratio}
\end{figure}

The tracking dataset, on the other hand, is composed of 20 videos. Each video is cut down into a variable number frames. For each video there is a corresponding \emph{.txt} file, where each line represents the bounding box data of the respective frame in the following format:
\[\mat{\text{id}_{\text{class}} & x_{\text{bbox-left}} & y_{\text{bbox-top}} & w_{\text{bbox}} &h_{\text{bbox}}} \]

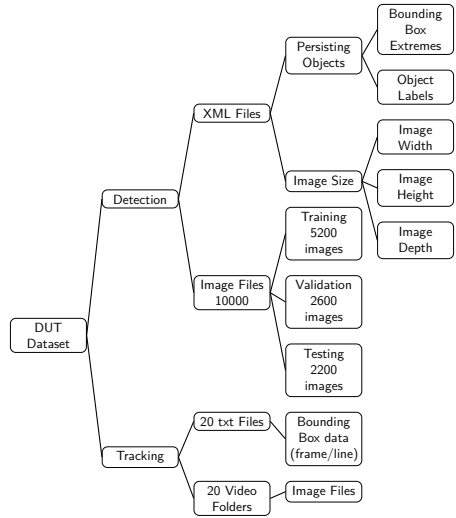
\begin{figure}[h]
    \centering
    \resizebox{0.4\textwidth}{!}{ 
        \begin{forest}
            for tree={
                font=\sffamily,
                grow=east,
                draw,
                rounded corners,
                node options={align=center, text width=5em},
                edge+={thick},
                parent anchor=east,
                child anchor=west,
                l sep=4mm,
                s sep=4mm
            }
            [DUT Dataset
            [
            Tracking
                [20 Video Folders
                    [Image Files]
                ]
                [20 txt Files
                    [Bounding Box data (frame/line)]
                ]
            ]
            [Detection
                [Image Files \\ 10000
                    [Testing \\ 2200 images]
                    [Validation \\ 2600 images]
                    [Training \\ 5200 images]                
                ]
                [XML Files
                    [Image Size
                        [Image Depth]
                        [Image Height]
                        [Image Width]
                    ]
                    [Persisting Objects
                        [Object Labels]
                        [Bounding Box Extremes]
                    ]
                ]
            ]
            ]
        \end{forest}
    }
    \caption{DUT Dataset Structure}
    \label{fig:DUT_Tree}
\end{figure}

\subsection{Detection Models}
In our study, all detection models requires the dataset to have a \emph{.text} file for each image, in which the ground truth classes of objects and their bounding boxes dimensions are reserved. A single object data should be written in the following format as a line in the \emph{.text} file: 
\[\mat{\text{id}_{\text{class}} & x_{\text{bbox-center}} & y_{\text{bbox-center}} &
w_{\text{bbox}} &h_{\text{bbox}}
}\]
For example, $\mat{1 & 100 & 150 & 50 & 30}$ means an object of class-id 1 is found inside a bounding box that is centered at $(100, 150)$, with a width of 50 and a height of 30, all in pixels.
Converting bounding boxes extremes to the models' compatible format could be done through the following equation:
\begin{equation}
    \begin{aligned}
        x_{\text{bbox-center}} &= \frac{x_{\text{min}} + x_{\text{max}}}{2} \\
        y_{\text{bbox-center}} &= \frac{y_{\text{min}} + y_{\text{max}}}{2} \\
        w_{\text{bbox}} &= \frac{x_{\text{min}} - x_{\text{max}}}{\text{ \small Image Width}}\\
        h_{\text{bbox}} &= \frac{y_{\text{min}} - y_{\text{max}}}{\text{ \small Image Height}}\\
    \end{aligned}
\end{equation}

\subsubsection*{Training and Validation}
For all models, the input image size was set to $640 \times 640$, with confidence threshold (\emph{conf}) = 0.001 during validation, Intersection over union threshold (\emph{IoU}) = 0.7, and with no drop outs. All models used CSPDarkNet53 backbone \cite{wang2020cspnet, bochkovskiy2020yolov4}, and Spatial Pyramid Pooling (SPPF) \cite{he2015spatial} in their neck. All models were trained for 50 epochs. Table \ref{tab:model_specs_and_losses} provides a comparison of model specifications. Figs. \ref{fig:val_batch} and \ref{fig:train_result} show a sample batch of validation data and training progress of YOLOv5x, respectively.

\begin{figure}[]
    \centering
    \includegraphics[width = \textwidth]{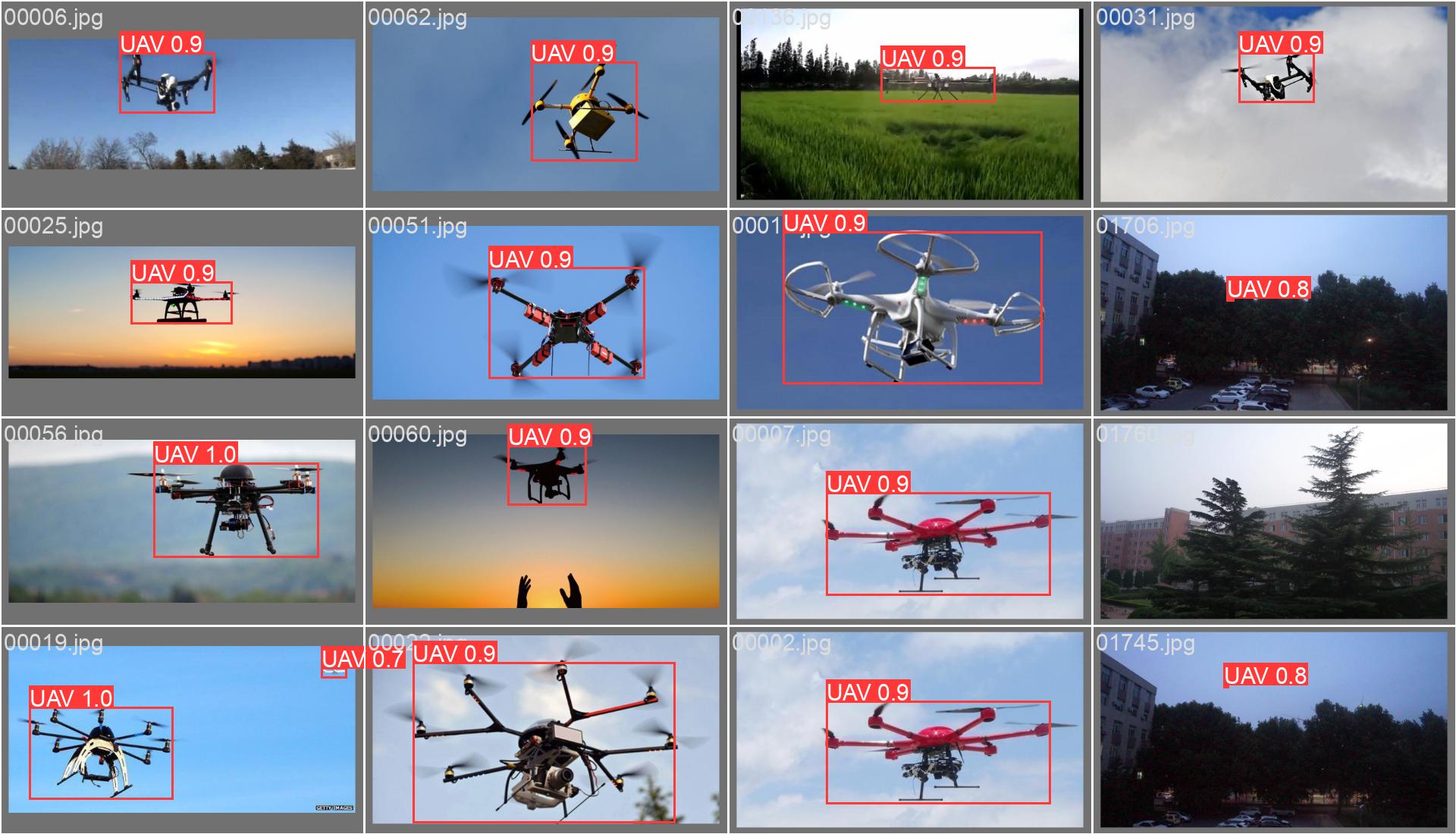}
    \caption{Sample Validation Batch.}
    \label{fig:val_batch}
\end{figure}

\begin{figure}[]
    \centering
    \includegraphics[width = \textwidth]{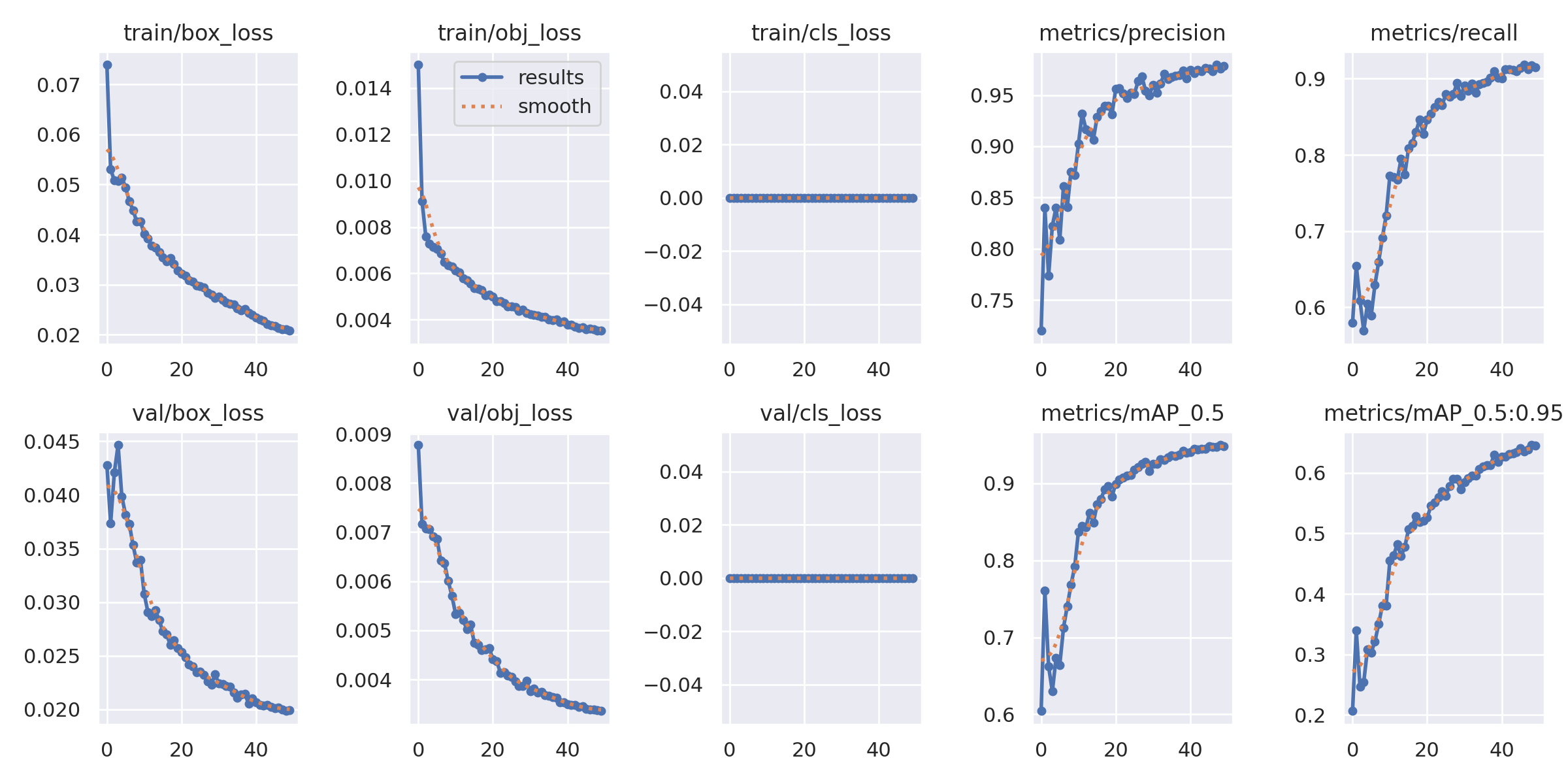}
    \caption{Sample Training Progress for YOLOv5x detection model.}
    \label{fig:train_result}
\end{figure}

\begin{table*}[h]
    \centering
    \caption{Comparison of Model Training Parameters}
    \label{tab:model_specs_and_losses}
    \begin{tabular}{@{}cccc@{}}
        \toprule
        Model & Trainable Parameters  & Batch Size \\
        \midrule
        YOLOv5n & \num[group-separator={,}]{1760518} & 64 \\
        YOLOv5s & \num[group-separator={,}]{7012822} & 64 \\
        YOLOv5l & \num[group-separator={,}]{46108278} & 16 \\
        YOLOv5x & \num[group-separator={,}]{86173414} & 16 \\
        YOLOv8n & \num[group-separator={,}]{3005843} & 64 \\
        YOLOv8s & \num[group-separator={,}]{11125971} & 64 \\
        YOLOv8l & \num[group-separator={,}]{43607379} & 16 \\
        YOLOv8x & \num[group-separator={,}]{68124531} & 16 \\
        \bottomrule
    \end{tabular}
\end{table*}

\subsection{Tracking Models}

When it comes to the tracking models, even though the ground truth bounding boxes representation is compatible with the object detection model's required format, the center of the bounding box is required for the tracker evaluation. Converting the bounding boxes from the ($x_{\text{left}}$, $y_{\text{top}}$, $width$, $height$) to ($x_{\text{center}}$, $y_{\text{center}}$, $width$, $height$) format could be done through the following equation (where the $w_{\text{bbox}}$ and $h_{\text{bbox}}$ remain unchanged):
\begin{equation}
    \begin{aligned}
        x_{\text{bbox-center}} &= x_{\text{left}} + \frac{w_{\text{bbox}}}{2} \\
        y_{\text{bbox-center}} &= y_{\text{top}} + \frac{h_{\text{bbox}}}{2} \\
    \end{aligned}
\end{equation}

\subsubsection{ByteTrack}
By associating a greater number of detection boxes, the technique presented in \cite{zhang2021bytetrack} seeks to enhance the performance of multi-object tracking (MOT). Conventional techniques simply take into account high-score detection boxes, which leaves out objects and causes trajectories to become fragmented. To solve this problem, the ByteTrack algorithm associates nearly all detection boxes—even the ones with low scores.

First, the algorithm recovers actual objects and filters out background detections by using the similarities between low score detection boxes and existing tracklets. It matches tracklets and detection boxes according to how similar their appearances or motions are. Tracklet locations in the next frame are predicted using a Kalman filter, and the similarity may be calculated using Re-ID feature distance or Intersection over Union (IoU).

DeepSORT and SORT algorithms are not as effective as byte track. Multi-object tracking accuracy for Bytetrack is 76.6 MOTA, whereas that of SORT and DeepSort is 74.6 and 75.4 MOTA, respectively \cite{zhang2021bytetrack}.

\subsubsection{BoT-SORT}
The Robust Associations Multi-Pedestrian Tracking (BoT-SORT) developed by \cite{aharon2022bot} is a modification of ByteTrack \cite{zhang2021bytetrack}, where it uses Kalman filters for modelling the object motion withing the image, enjoys corrections of the object state to compensate the camera motion, and fuses the Intersection-over-Union (\emph{IoU}) with the re-identification (\emph{Re-ID}), i.e. matching the object features across frames, as a tracking metric. Figure \ref{fig:botsortflow} shows the BoT-SORT algorithm flow as provided by \cite{aharon2022bot}.

\begin{figure}[H]
    \centering
    \includegraphics[scale = 0.3]{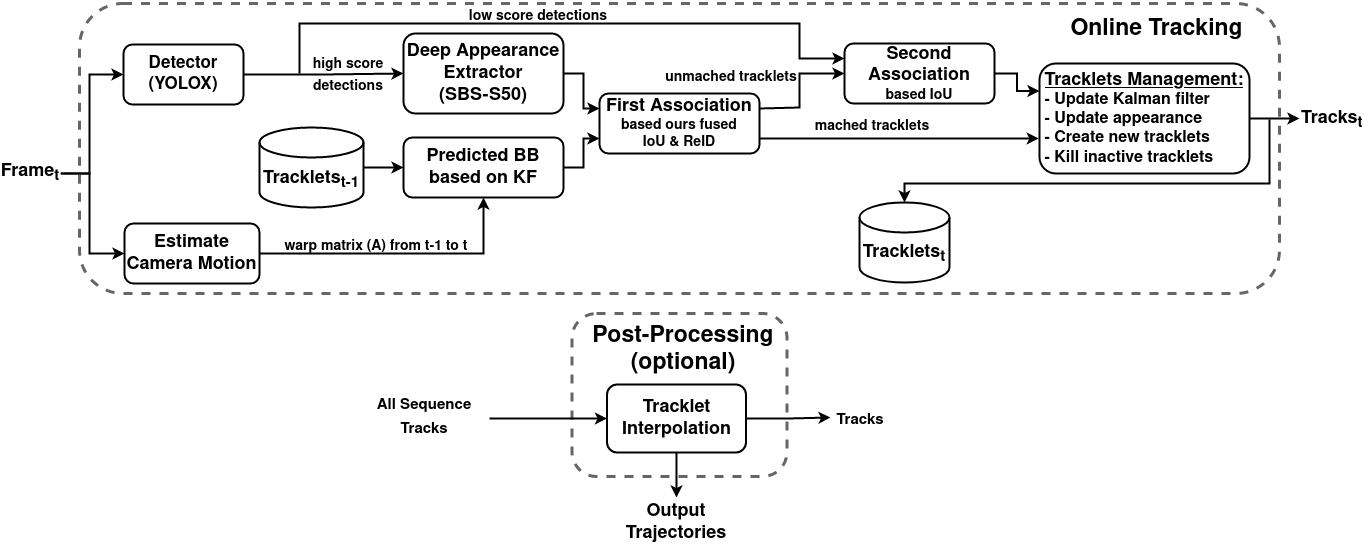}
    \caption{BoT-SORT-ReID tracker pipeline (retrieved from \cite{aharon2022bot})}
    \label{fig:botsortflow}
\end{figure}

\subsection{Computational Resources}
The data generation, detectors training, detectors inference, and tracking were all done using Google Colab's Nvidia A100-40GB GPU. The codes were based on Python and PyTorch v1.12? 

\subsection{Evaluation Metrics}
\label{subsec:evaluation-metrics}
\subsubsection{Intersection over Union ($IoU$)}

It is the ratio between the area of overlap between the predicted and the ground truth bounding boxes to the area of their union.
$\frac{\textit{Area of Overlap}}{\textit{Area of Union}}$
There are 0 to 1 IoU scores, with 1 denoting perfect alignment between the ground truth and forecast boxes. A common threshold, like 0.5, is frequently employed in practical applications to assess if a detection is a true positive. Stricter evaluation requirements brought about by higher 
$IoU$ thresholds make it more difficult for a detection to be considered accurate.

\subsubsection{Mean Average Precision ($mAP$)}

Average Precision is the area under the precision-recall curve. The mean Average Precision is the just avergaing those areas across all classes. $mAP$ is used with a certain $IoU$ threshold. $mAP50$ uses the curves plotted with $IoU$ = 0.5, while $mAP50-90$ is calculated from precision-recall curves plotted with IoU thresholds from 0.5 to 0.95 in steps of 0.05. A high $mAP50$ score indicates that the detector performs well at recognizing objects with a moderate overlap (50\%) between the predicted and ground truth boxes. In order to function successfully at both more lenient (lower overlap) and higher IoU criteria (closer to perfect overlap), the model must be both robust and precise. Compared to $mAP50$, $mAP50-95$ is thought to be a more thorough and rigorous evaluation metric since it takes into consideration a larger variety of detection circumstances, providing a more accurate overall evaluation of the model's performance in practical applications.

\section{Results and Discussions}
\label{sec:results-discussions}

In this section, we present the results and discussions based on the extensive performance evaluations carried out. First, the analysis of detection models' performance is presented. Second, the results of selected tracking methods are analysed. Third, a detailed discussion is provided based on the presented results.

\subsection{Detection}

All models were evaluated on the testing dataset, using \emph{mAP50},\emph{ mAP50-95}, \emph{Precision}, and \emph{Recall} as metrics for both validation and testing phases. Table \ref{tab:val_test_perform} presents a comparison of validation and test performances across all models. YOLOv5x outperforms all models, and YOLOv5 models in general are better than YOLOv8's. Figure \ref{fig:model_comparison} shows that the previous statement is evident. However, YOLOv-x models showed more ability to detect unrecognizable objects, such as in blurred images (see Fig. \ref{fig:model_comparison} (i, j, k, and l)). This could be explained by relatively higher number of model parameters; which gives more complexity to the YOLOv-x models. Overall, the models are capable of extracting the meaningful features for UAVs. In fact, the models were able to detect the shadow of a UAV as a UAV (see Fig. \ref{fig:model_comparison} (m, n, o , and p)). It was noted that for YOLOv5, as the model gets more complex, the performance as well gets enhanced. On the other hand, that was not the case for the YOLOv8 model. This could be attributed to the number of epochs used in such comparison.

\begin{table}[H]
    \centering
    \caption{Comparison of Validation and Test Performances across all Models}
    \label{tab:val_test_perform}
    \begin{tabular}{@{}c|c|c|cc|cc|cc|cc@{}}
        \toprule
        Model & Model & Inference & \multicolumn{2}{c}{mAP50} & \multicolumn{2}{c}{mAP50--95} & \multicolumn{2}{c}{Precision} & \multicolumn{2}{c}{Recall} \\
        & Structure & & V & T & V & T & V & T & V & T \\
        \midrule
        YOLOv5 & Nano & \textbf{0.4ms} & 0.878 & 0.927 & 0.533 & 0.597 & 0.953 & 0.942 & 0.821 & 0.891 \\
        YOLOv5 & Small & 0.7ms & 0.925 & 0.954 & 0.590 & 0.643 & 0.956 & 0.969 & 0.887 & 0.925 \\
        YOLOv5 & Large & 2.3ms & \textbf{0.946} & 0.965 & {0.632} & 0.693 & \textbf{0.977} & 0.969 & 0.903 & 0.95 \\
        YOLOv5 & X & 3.2ms & 0.95 & \textbf{0.976} & \textbf{0.647} & \textbf{0.705} & {0.976} & 0.976 & \textbf{0.918} & \textbf{0.954} \\
        \midrule
        YOLOv8 & Nano & 0.5ms & 0.846 & 0.908 & 0.530 & 0.613 & 0.928 & 0.925 & 0.773 & 0.856 \\
        YOLOv8 & Small & 0.83ms & 0.921 & 0.872 & 0.562 & 0.643 & 0.941 & 0.94 & 0.784 & 0.871 \\
        YOLOv8 & Large & 2.5ms & 0.854 & 0.913 & 0.553 & 0.634 & 0.924 & 0.934 & 0.774 & 0.847 \\
        YOLOv8 & X & 2.66ms & 0.845 & 0.904 & 0.555 & 0.632 & 0.921 & 0.926 & 0.764 & 0.848 \\
        \bottomrule
    \end{tabular}
\end{table}

\begin{figure}[]
    \centering
    \begin{subfigure}{0.225\textwidth}
        \centering
        \includegraphics[scale=0.06]{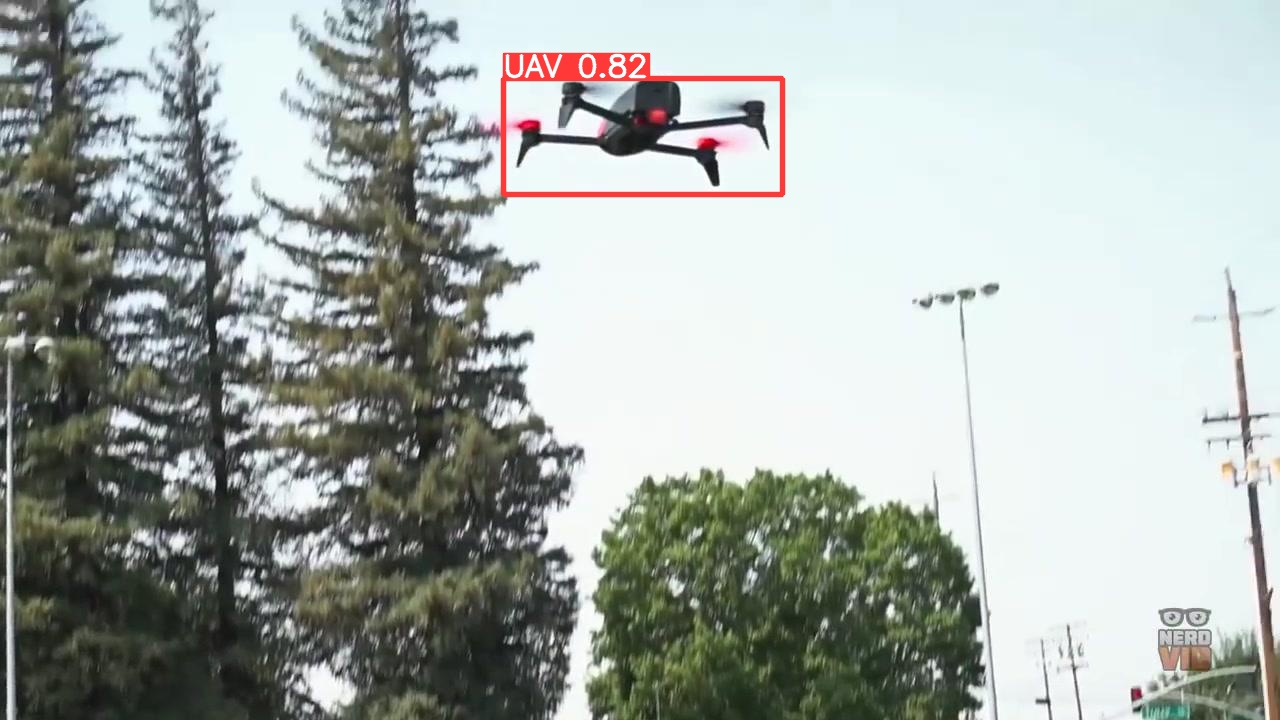}
        \caption{YOLOv5s - Image 1}
        \label{fig:yolov5s1}
    \end{subfigure}
    \begin{subfigure}{0.225\textwidth}
        \centering
        \includegraphics[scale=0.06]{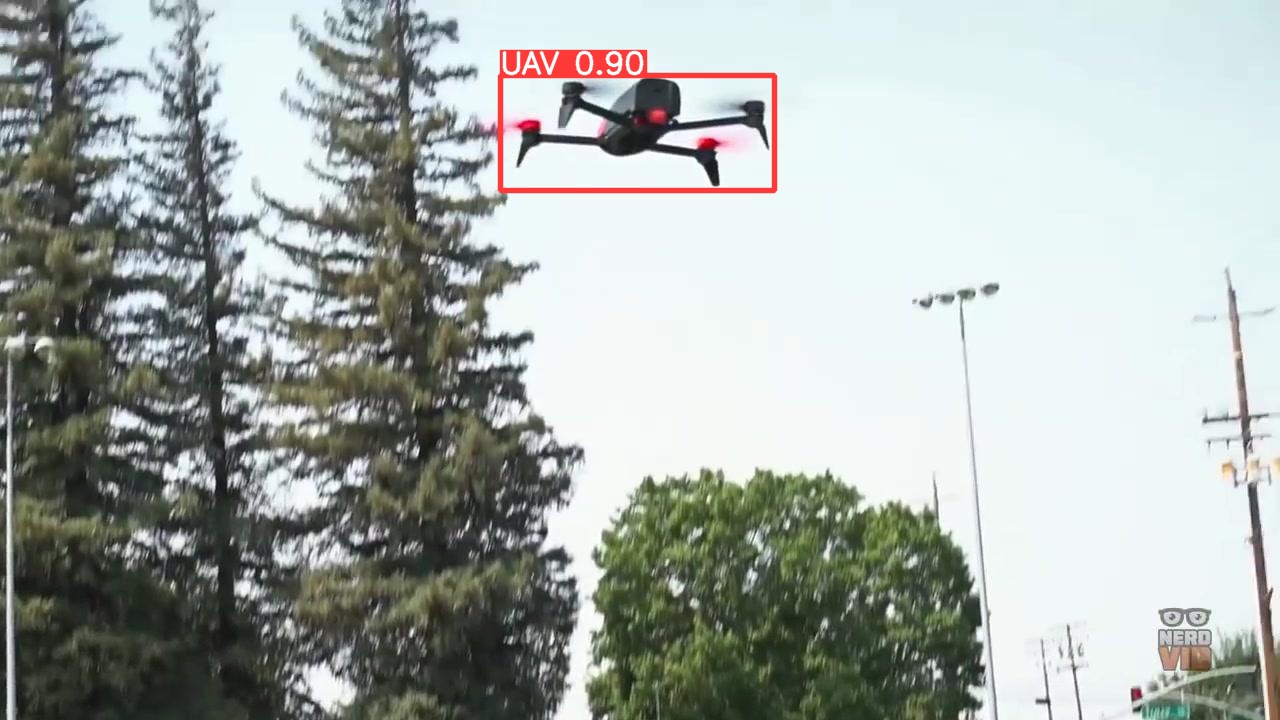}
        \caption{YOLOv5x - Image 1}
        \label{fig:yolov5x1}
    \end{subfigure}
    \begin{subfigure}{0.225\textwidth}
        \centering
        \includegraphics[scale=0.06]{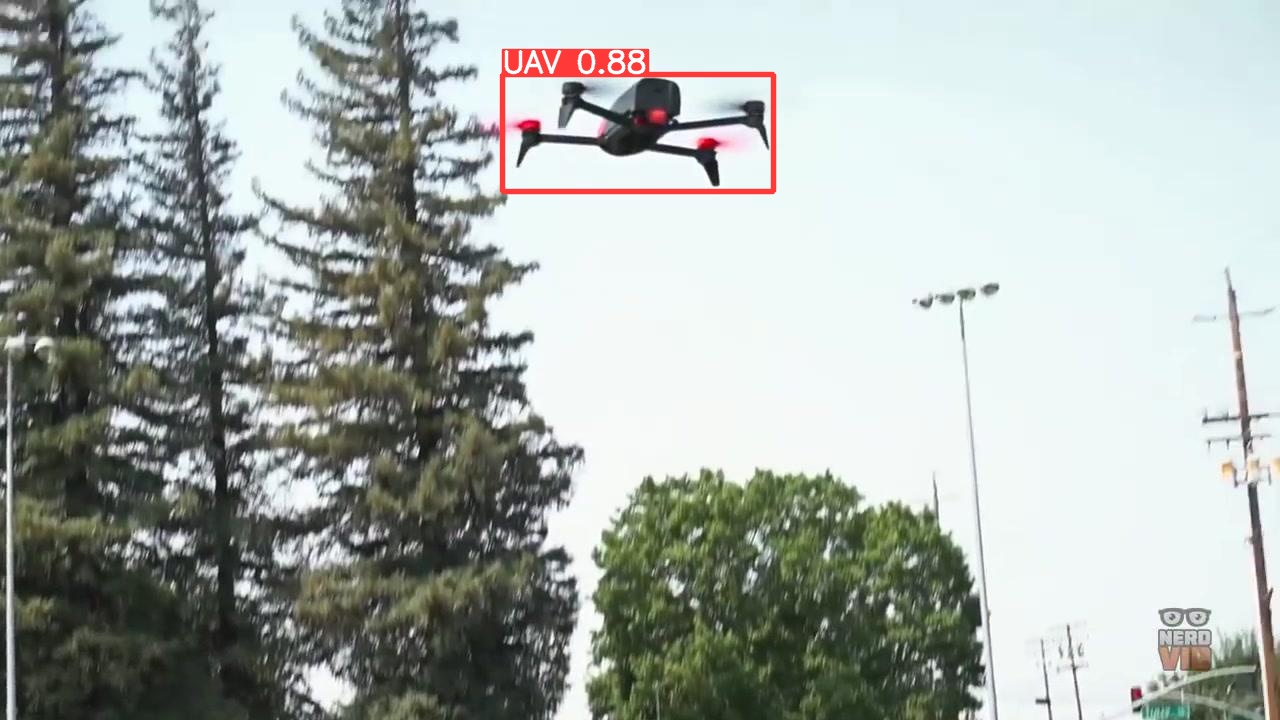}
        \caption{YOLOv8s - Image 1}
        \label{fig:yolov8s1}
    \end{subfigure}
    \begin{subfigure}{0.225\textwidth}
        \centering
        \includegraphics[scale=0.06]{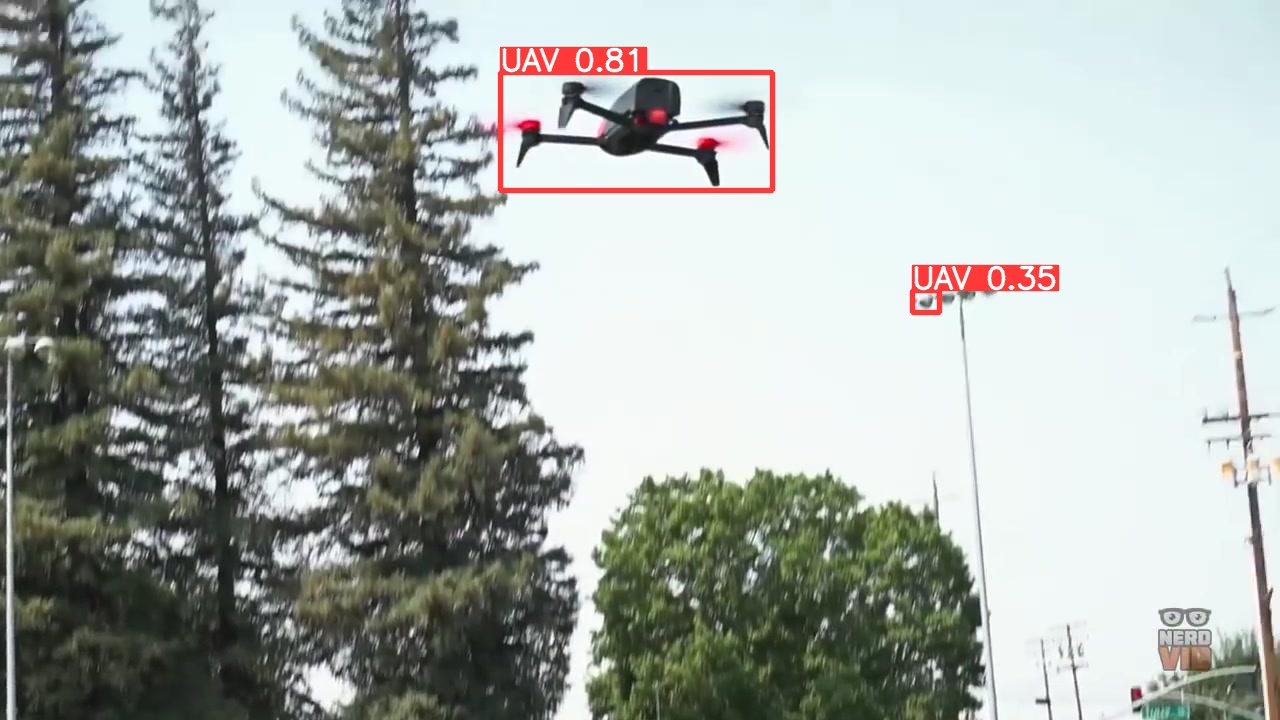}
        \caption{YOLOv8x - Image 1}
        \label{fig:yolov8x1}
    \end{subfigure}
    \\
    \begin{subfigure}{0.225\textwidth}
        \centering
        \includegraphics[scale=0.06]{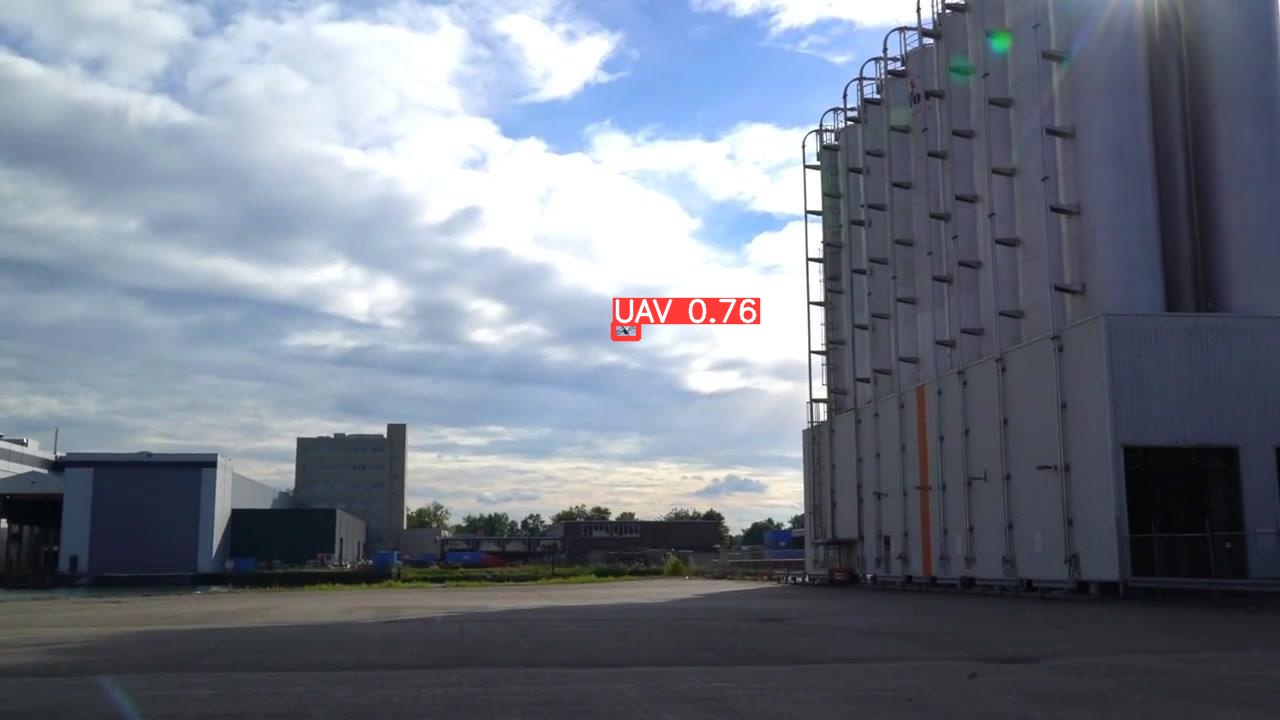}
        \caption{YOLOv5s - Image 2}
        \label{fig:yolov5s2}
    \end{subfigure}
    \begin{subfigure}{0.225\textwidth}
        \centering
        \includegraphics[scale=0.06]{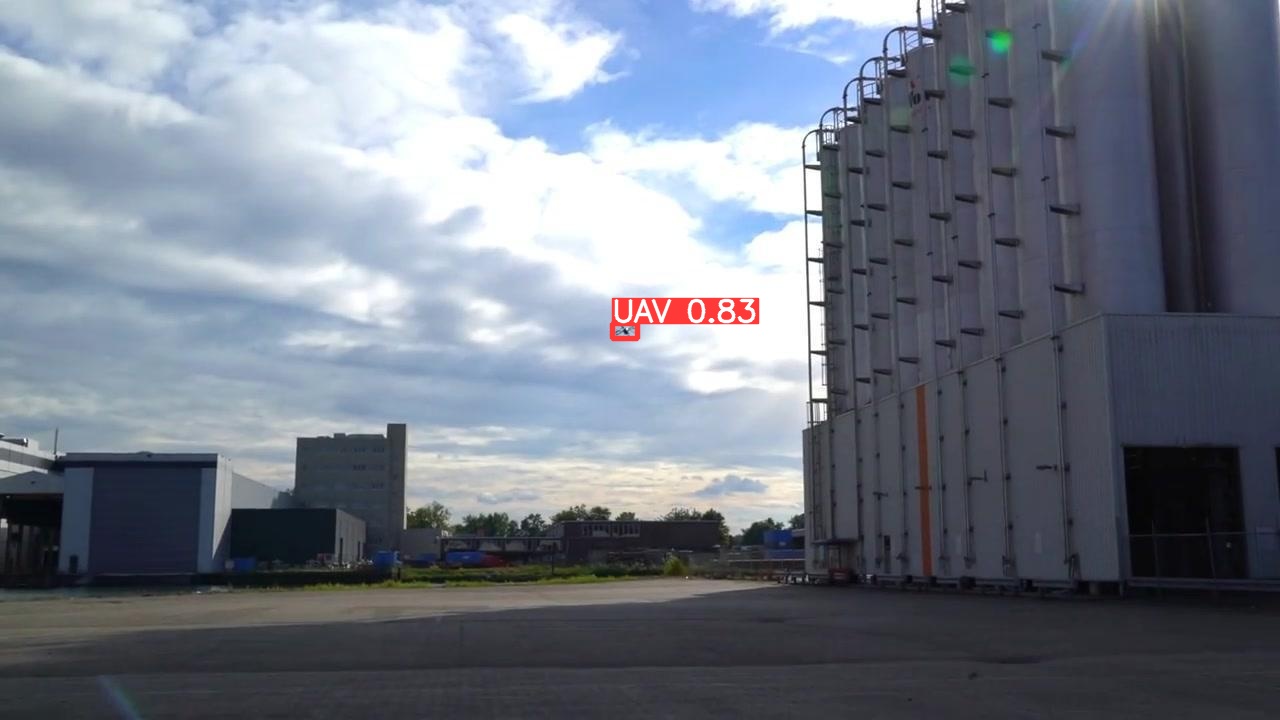}
        \caption{YOLOv5x - Image 2}
        \label{fig:yolov5x2}
    \end{subfigure}
    \begin{subfigure}{0.225\textwidth}
        \centering
        \includegraphics[scale=0.06]{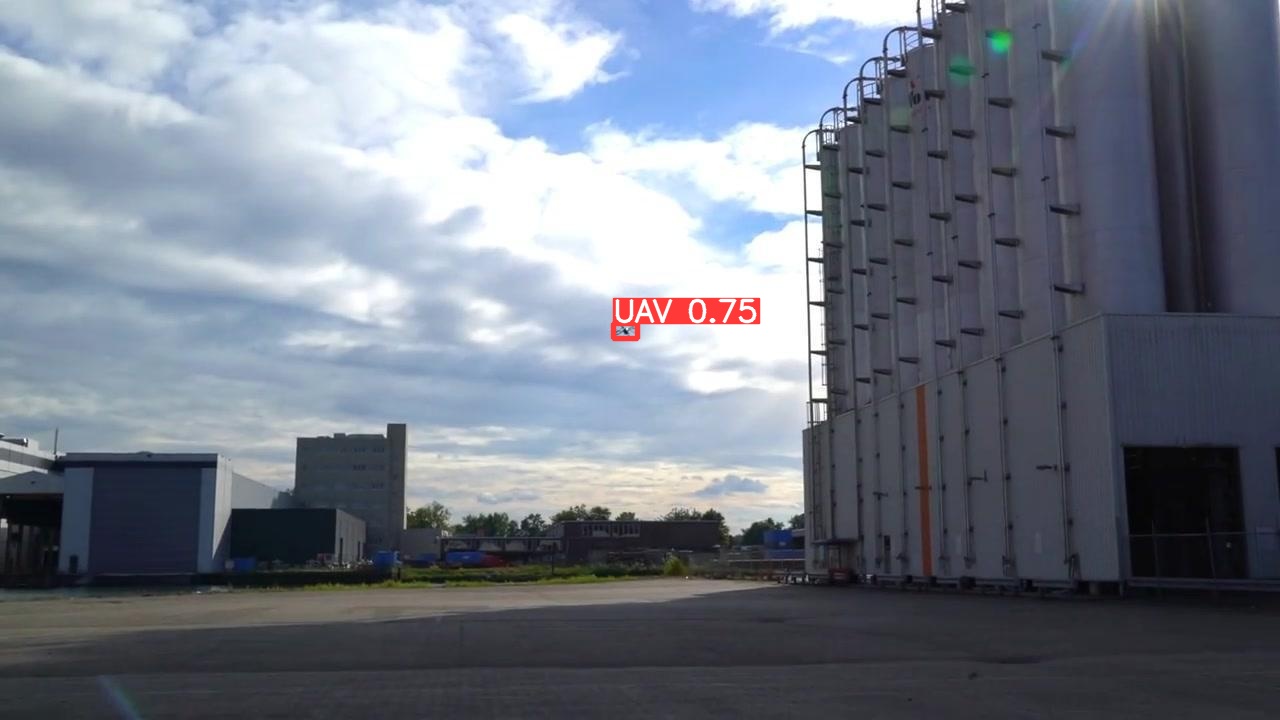}
        \caption{YOLOv8s - Image 2}
        \label{fig:yolov8s2}
    \end{subfigure}
    \begin{subfigure}{0.225\textwidth}
        \centering
        \includegraphics[scale=0.06]{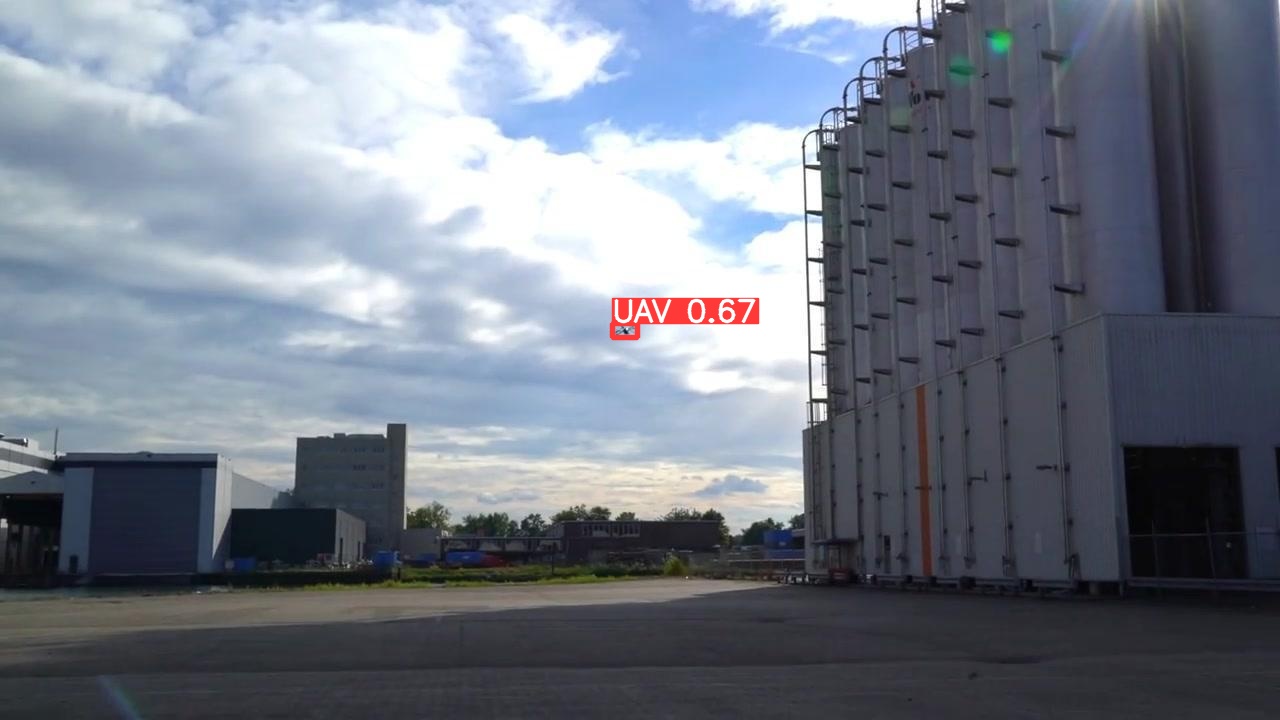}
        \caption{YOLOv8x - Image 2}
        \label{fig:yolov8x2}
    \end{subfigure}
    \\
    \begin{subfigure}{0.225\textwidth}
        \centering
        \includegraphics[scale=0.06]{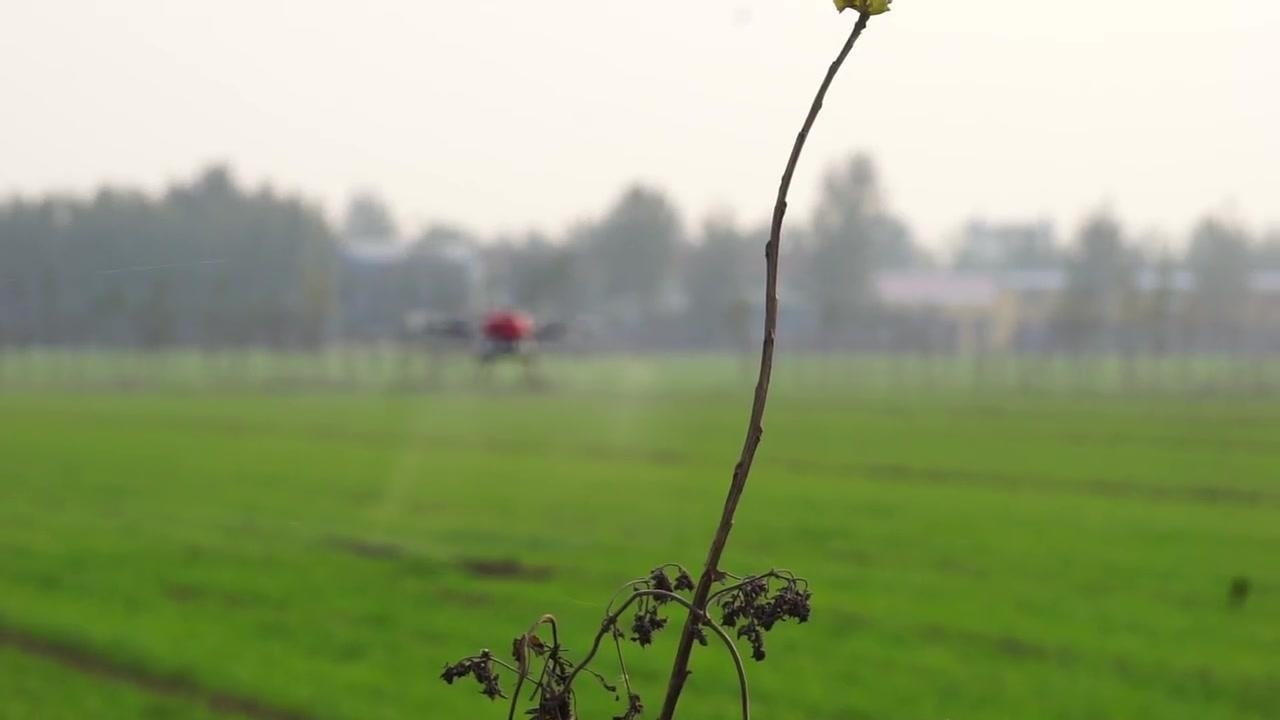}
        \caption{YOLOv5s - Image 3}
        \label{fig:yolov5s3}
    \end{subfigure}
    \begin{subfigure}{0.225\textwidth}
        \centering
        \includegraphics[scale=0.06]{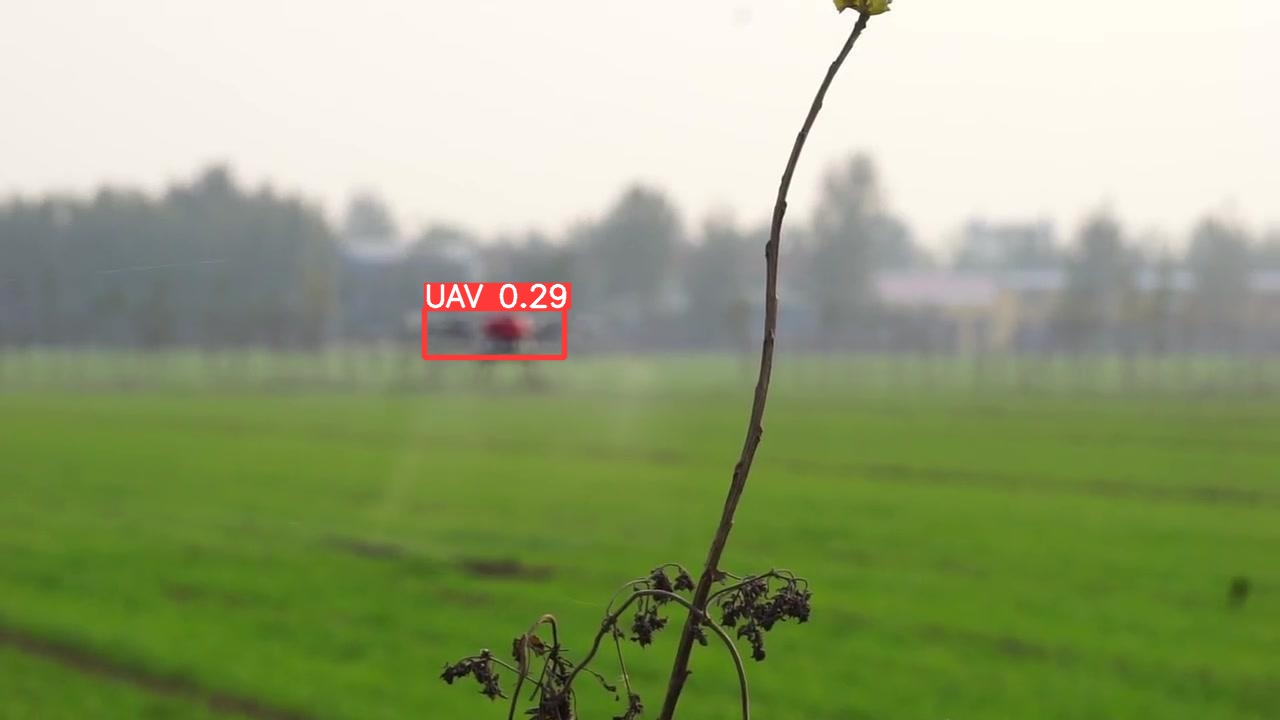}
        \caption{YOLOv5x - Image 3}
        \label{fig:yolov5x3}
    \end{subfigure}
    \begin{subfigure}{0.225\textwidth}
        \centering
        \includegraphics[scale=0.06]{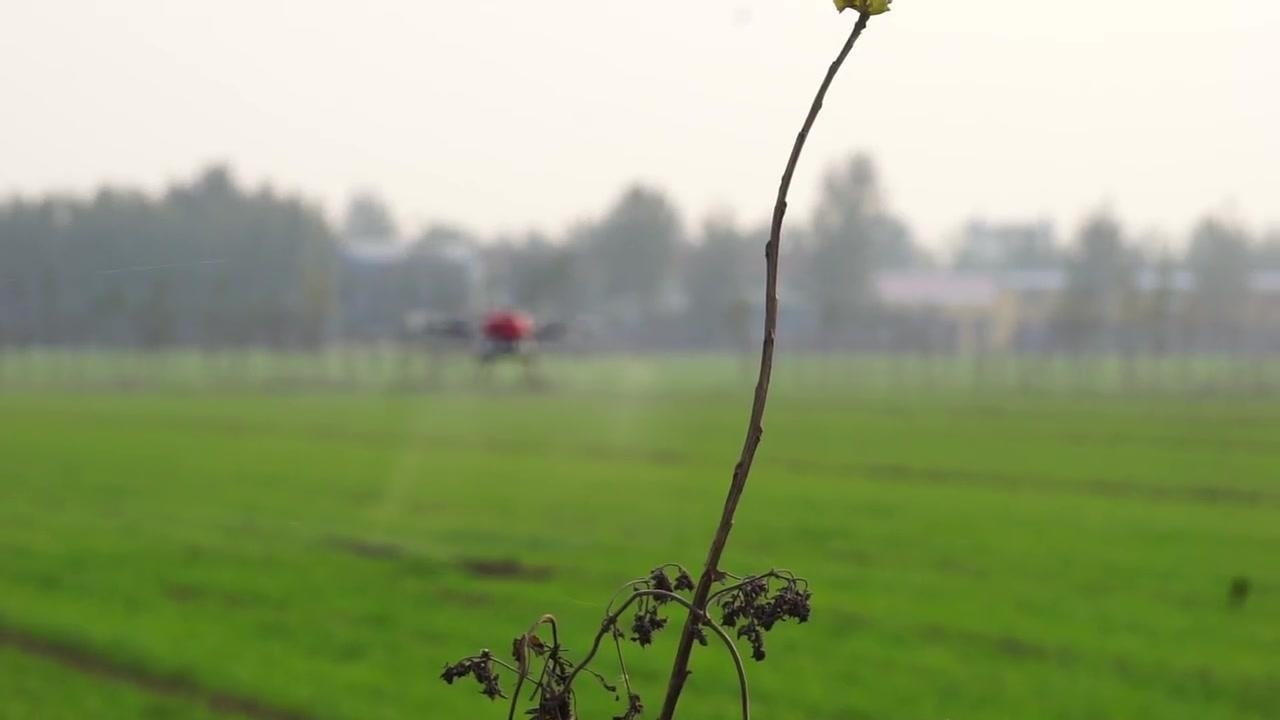}
        \caption{YOLOv8s - Image 3}
        \label{fig:yolov8s3}
    \end{subfigure}
    \begin{subfigure}{0.225\textwidth}
        \centering
        \includegraphics[scale=0.06]{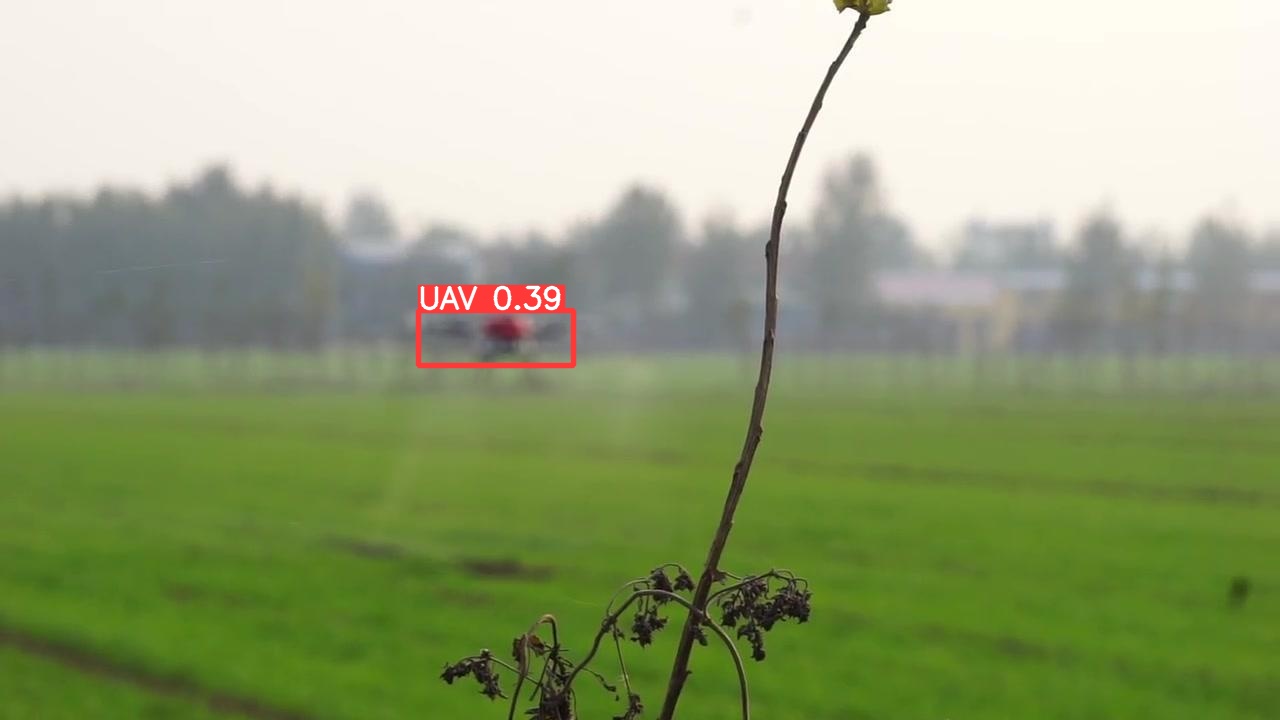}
        \caption{YOLOv8x - Image 3}
        \label{fig:yolov8x3}
    \end{subfigure}
    \\
    \begin{subfigure}{0.225\textwidth}
        \centering
        \includegraphics[scale=0.06]{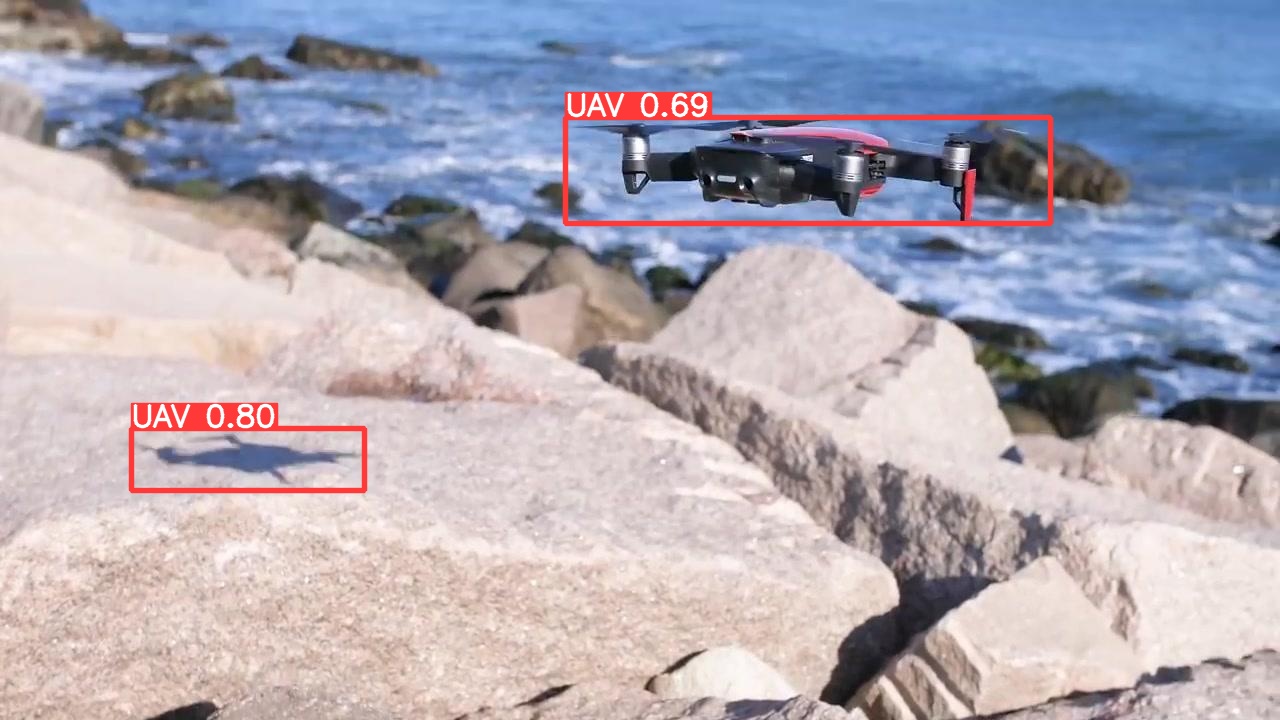}
        \caption{YOLOv5s - Image 4}
        \label{fig:yolov5s4}
    \end{subfigure}
    \begin{subfigure}{0.225\textwidth}
        \centering
        \includegraphics[scale=0.06]{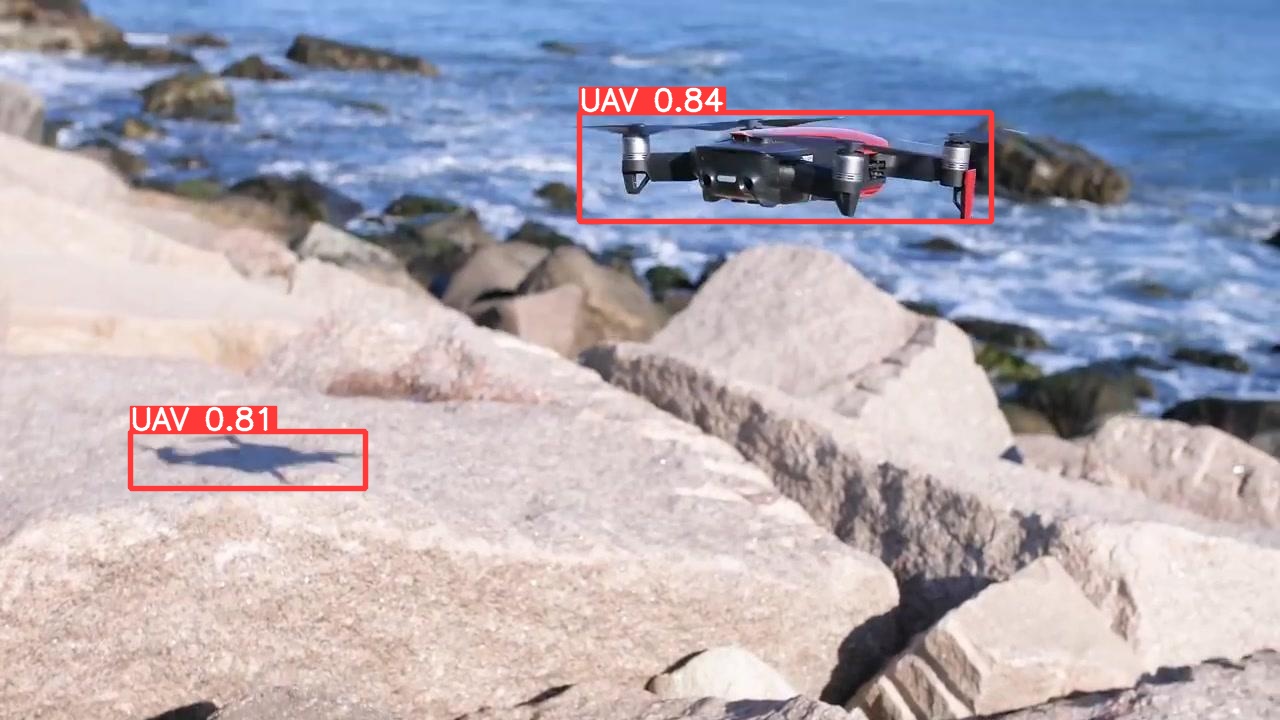}
        \caption{YOLOv5x - Image 4}
        \label{fig:yolov5x4}
    \end{subfigure}
    \begin{subfigure}{0.225\textwidth}
        \centering
        \includegraphics[scale=0.06]{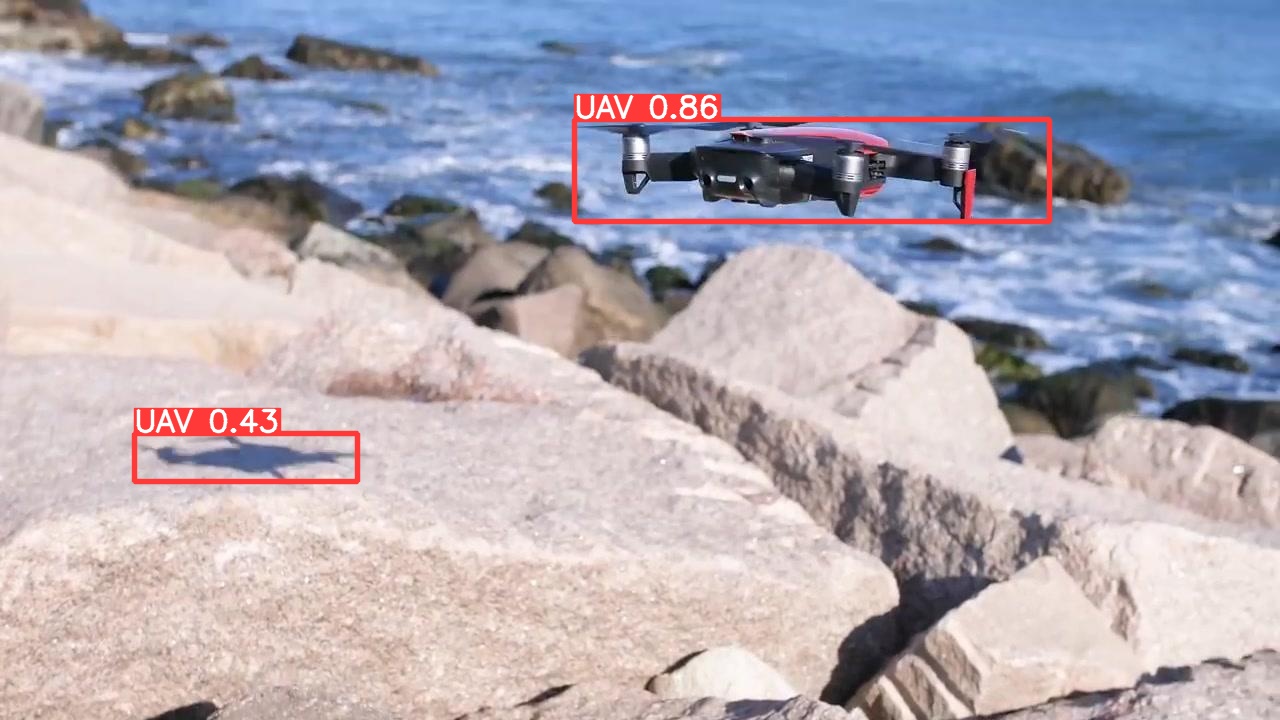}
        \caption{YOLOv8s - Image 4}
        \label{fig:yolov8s4}
    \end{subfigure}
    \begin{subfigure}{0.225\textwidth}
        \centering
        \includegraphics[scale=0.06]{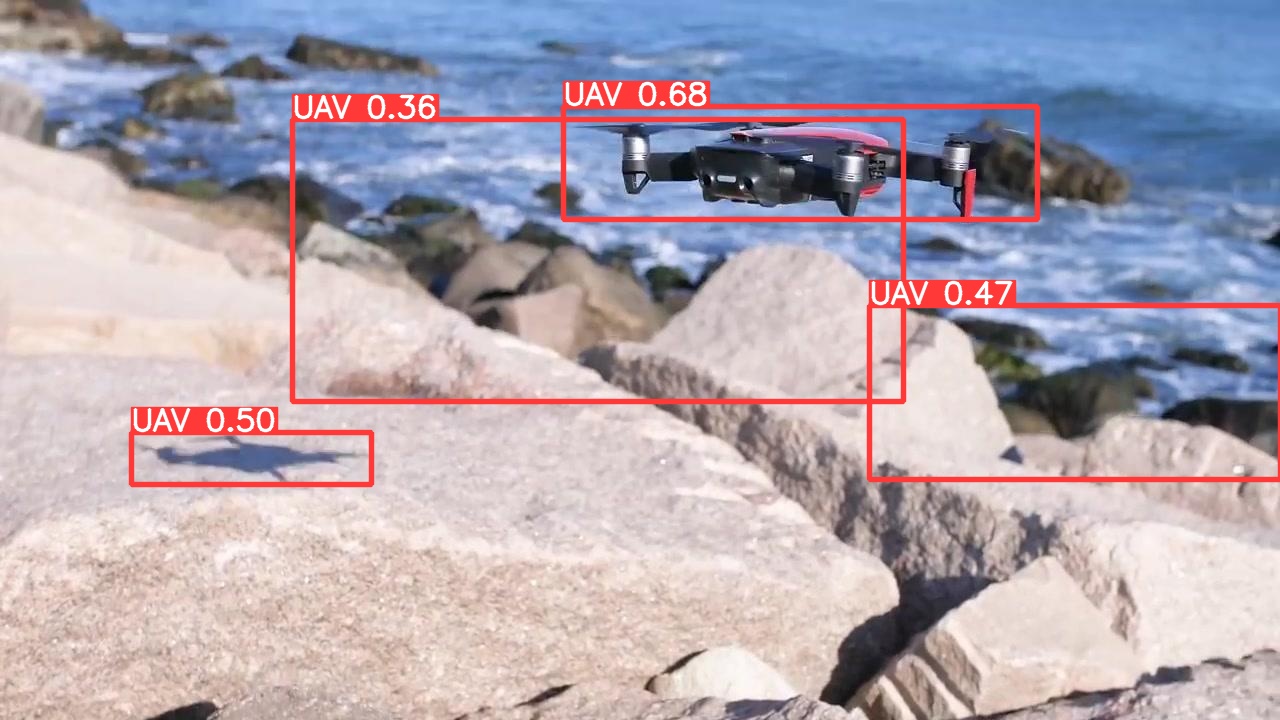}
        \caption{YOLOv8x - Image 4}
        \label{fig:yolov8x4}
    \end{subfigure}
    \caption{Comparison of model outputs for YOLOv5s, YOLOv5x, YOLOv8s, and YOLOv8x}
    \label{fig:model_comparison}
\end{figure}

\begin{figure}[]
    \centering
    \includegraphics[width =\textwidth]{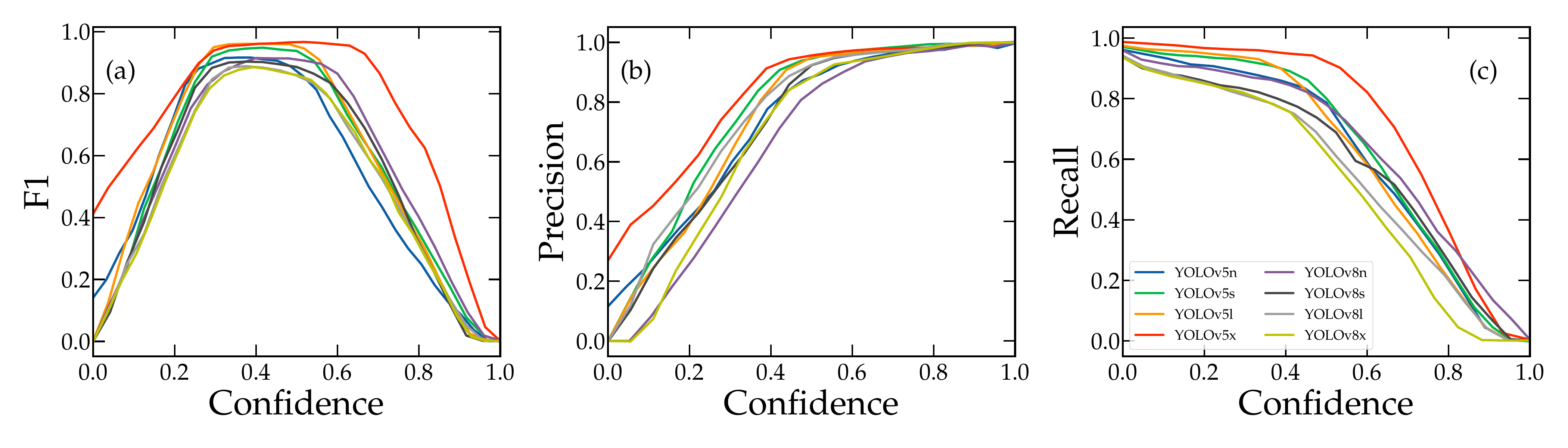}
    \caption{Performance curves (F1, Precision, and Recall metrics) for each model at varying confidence thresholds.}
    \label{fig:yolo_comp}
\end{figure}

Further assessments were done to check how the least performing model would perform on images with confusing objects. YOLOv8x was tested on a set of 20 challenging images that featured UAVs and birds with similar colors and overlapping elements (\ref{tab:yolov8xconfusing}. Images in Fig. \ref{fig:confusing} shows that even though YOLOv8x is relatively a low performer, its performance is accepted. Thus, YOLOv8x is used in the tracker for further assessment.

\begin{table}[h]
    \centering
    \caption{Results summary of YOLOv8x Model on the confusing/challenging test dataset}
    \label{tab:yolov8xconfusing}
    \begin{tabular}{@{}c|c|c|c|c@{}}
        \toprule
        Model & mAP50 & mAP50--95 & Precision & Recall \\
        \midrule
        YOLOv8x & 0.5747 & 0.2158 & 0.7717 & 0.4667 \\
        \bottomrule
    \end{tabular}
\end{table}

\begin{figure}[H]
    \centering
    \includegraphics[scale = 1]{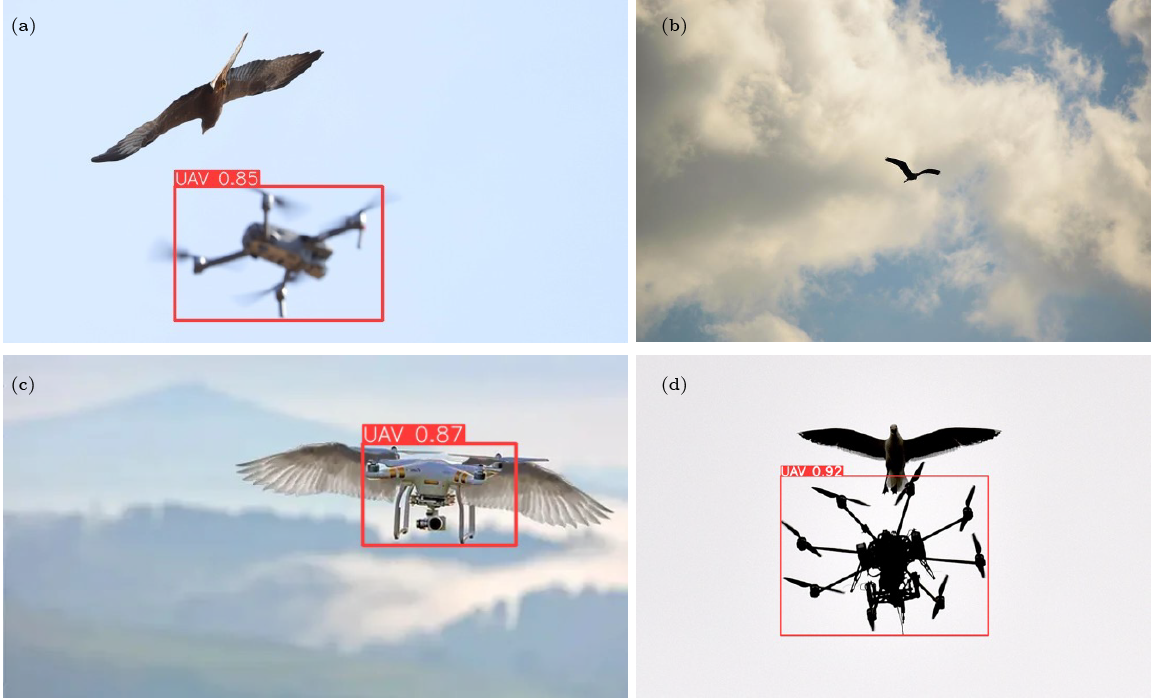}
    \caption{Examples of UAV detection in confusing scenarios. (a) A UAV detected with a confidence score of 0.85 despite being partially obscured by a bird. (b) Correct mislabeling by the model showcasing the model's challenge in distinguishing between UAVs and birds. (c) A UAV detected with a confidence score of 0.87, where the wings of a bird and the body of the UAV are closely overlapping, illustrating the model's ability to identify UAVs in complex visual overlaps. (d) A UAV detected with a high confidence score of 0.92, even though it is almost entirely blended with a bird.}
    \label{fig:confusing}
\end{figure}

\begin{table}[]
    \centering
    \caption{Comparison of Botsort and Byte Track Performances}
    \label{tab:botsort_bytetrack_perform}
    \begin{tabular}{@{}cc|cc|cc@{}}
        \toprule
        \multicolumn{2}{c|}{Video} & \multicolumn{2}{c|}{Botsort} & \multicolumn{2}{c}{Byte Track} \\
         Video \# & Sequence Length & Mean IoU & Mean Center Error (pixels) & Mean IoU & Mean Center Error (pixels)\\
        \midrule
        1 & 1050 & \textbf{0.8497} & \textbf{4.6925} & 0.8407 & 5.2878 \\
        2 & 83 & \textbf{0.7337} & \textbf{2.1264} & 0.7332 & 2.1546 \\
        3 & 100 & 0.8521 & \textbf{1.5706} & \textbf{0.8522} & 1.6189 \\
        4 & 341 & \textbf{0.8076} & \textbf{6.2901} & 0.7351 & 7.5628 \\
        5 & 750 & \textbf{0.7971} & 3.4217 & 0.7878 & \textbf{3.4017} \\
        6 & 200 & 0.9032 & \textbf{2.1953} & \textbf{0.9036} & 2.2401 \\
        7 & 2480 & \textbf{0.8663} & \textbf{4.1938} & 0.8349 & 5.5345 \\
        8 & 2305 & \textbf{0.8567} & \textbf{2.7463} & 0.8365 & 3.5085 \\
        9 & 2500 & \textbf{0.9084} & \textbf{2.5025} & 0.9018 & 2.5318 \\
        10 & 2635 & \textbf{0.8425} & \textbf{3.7746} & 0.8416 & 3.7763 \\
        11 & 1000 & \textbf{0.8133} & \textbf{3.8124} & 0.7990 & 3.8376 \\
        12 & 1485 & \textbf{0.6248} & \textbf{2.6280} & 0.6192 & 2.6862 \\
        13 & 1915 & 0.5747 & 2.7177 & \textbf{0.5763} & \textbf{2.7113} \\
        14 & 590 & \textbf{0.6903} & 3.8453 & 0.6859 & \textbf{3.8443} \\
        15 & 1350 & \textbf{0.6893} & 3.1977 & 0.6716 & \textbf{3.1801} \\
        16 & 1285 & \textbf{0.6392} & \textbf{2.9583} & 0.6356 & 2.9668 \\
        17 & 780 & \textbf{0.5929} & \textbf{3.8938} & 0.5923 & 3.8989 \\
        18 & 1320 & \textbf{0.6749} & \textbf{1.9784} & 0.6715 & 1.9883 \\
        19 & 1300 & \textbf{0.6269} & \textbf{2.3412} & 0.6255 & 2.3463 \\
        20 & 1635 & \textbf{0.7222} & 2.8709 & 0.7085 & \textbf{2.8563} \\
        \bottomrule
    \end{tabular}
\end{table}

\subsection{Tracking}

The evaluation of tracking performance in this study utilizes two key metrics: Mean IoU (Intersection over Union) and Mean Center Error. Mean IoU measures the overlap between the predicted bounding box and the ground truth bounding box, providing an indication of how accurately the tracker detects the object's position and size. Higher IoU values indicate better performance. Mean Center Error calculates the average distance in pixels between the predicted and ground truth center points of the bounding boxes, reflecting the precision of the tracker in locating the object. Lower center error values signify more accurate tracking.

Table \ref{tab:botsort_bytetrack_perform} compares the performance of Botsort and Byte Track across 20 different videos, using two metrics: Mean IoU (Intersection over Union) and Mean Center Error (in pixels). Botsort outperforms Byte Track in terms of Mean IoU in 18 out of 20 videos. The largest difference in IoU is observed in video 4, where Botsort has an IoU of 0.8076 compared to Byte Track's 0.7351. In terms of Mean Center Error, Botsort performs better in 16 out of 20 videos. The smallest center error for Botsort is 1.5706 in video 3, which is also the smallest error across both methods. The largest difference in center error is in video 7, with Botsort having an error of 4.1938 and Byte Track having 5.5345. Overall, Botsort demonstrates superior performance in both IoU and center error metrics across the majority of the videos.

\begin{figure}[]
    \centering
    \includegraphics[scale = 0.23]{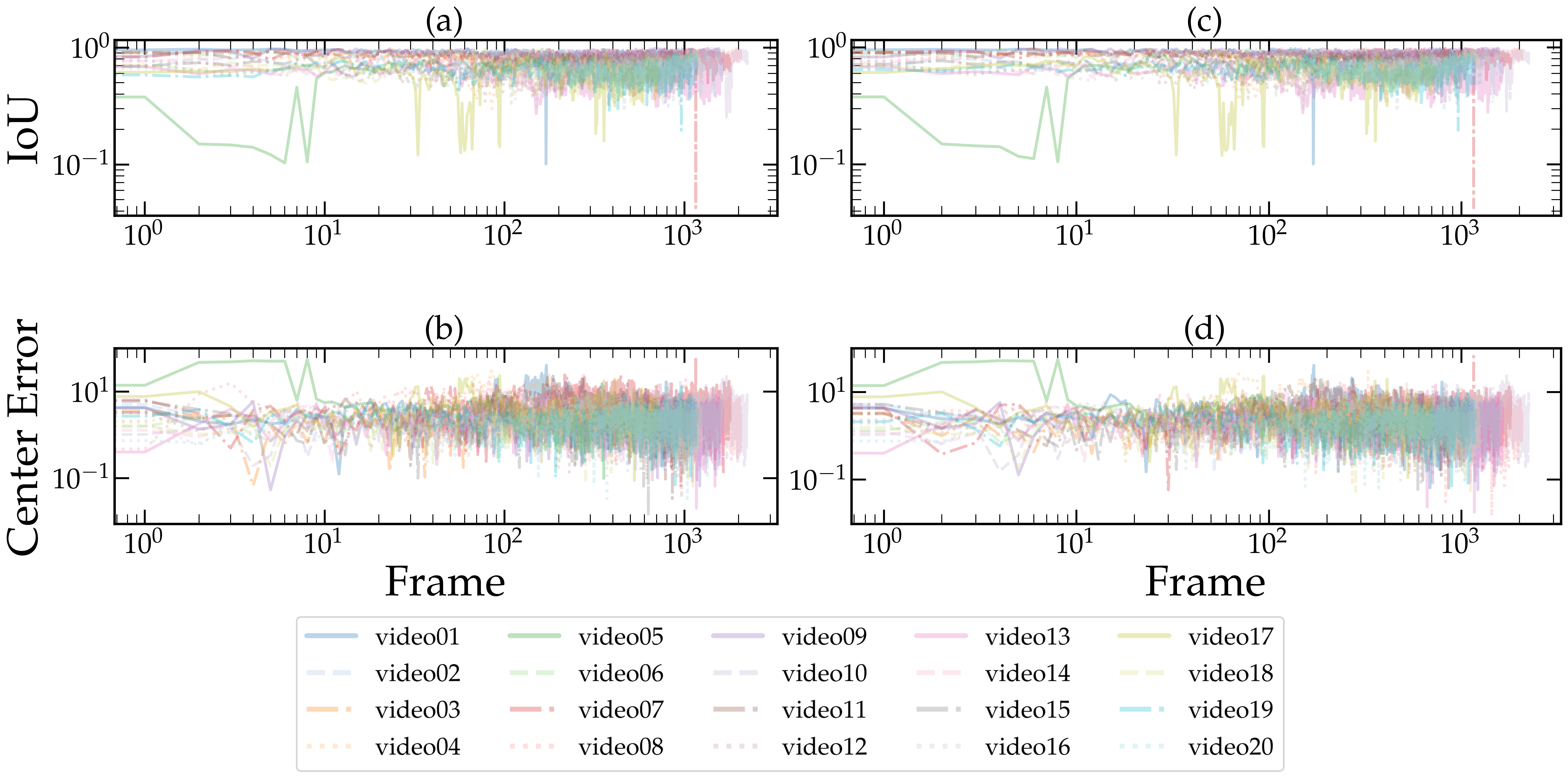}
    \caption{Tracking performance evaluation of Byte track and Botsort tracker methods. (a) Byte track's IoU and (b) Byte track's Center error. Moreover, (c) Bot sort's IoU and finally (d) Bot sort's Center error in pixels. They are plotted in log-scale in order to give more for visibility and to avoid clutter. Color-coded lines represent different videos (video \#1 to video \# 20). Frames on the x-axis depict the tracking performance over time. Figures might look the same but there exist minor variations in both trackers' performance.}
    \label{fig:trackers_comp}
\end{figure}

Figure \ref{fig:trackers_comp} consists of four plots, displaying the performance metrics for Botsort and Byte Track: (a) and (c) show IoU over frames for Botsort and Byte Track, respectively, while (b) and (d) show Center Error over frames for Botsort and Byte Track, respectively. Both methods show a high IoU (close to 1) for the majority of frames, with occasional drops indicating possible tracking errors or occlusions. Botsort (a) appears slightly more stable than Byte Track (c), with fewer significant drops in IoU. In terms of Center Error, both methods show errors fluctuating around a low value, typically below 10 pixels. Some videos exhibit higher center errors, suggesting more challenging tracking conditions. Botsort (b) again seems to maintain a more consistent performance with fewer spikes compared to Byte Track (d).

\begin{figure}[]
    \centering
    \begin{subfigure}{0.2\textwidth}
        \centering
        \includegraphics[scale=0.06]{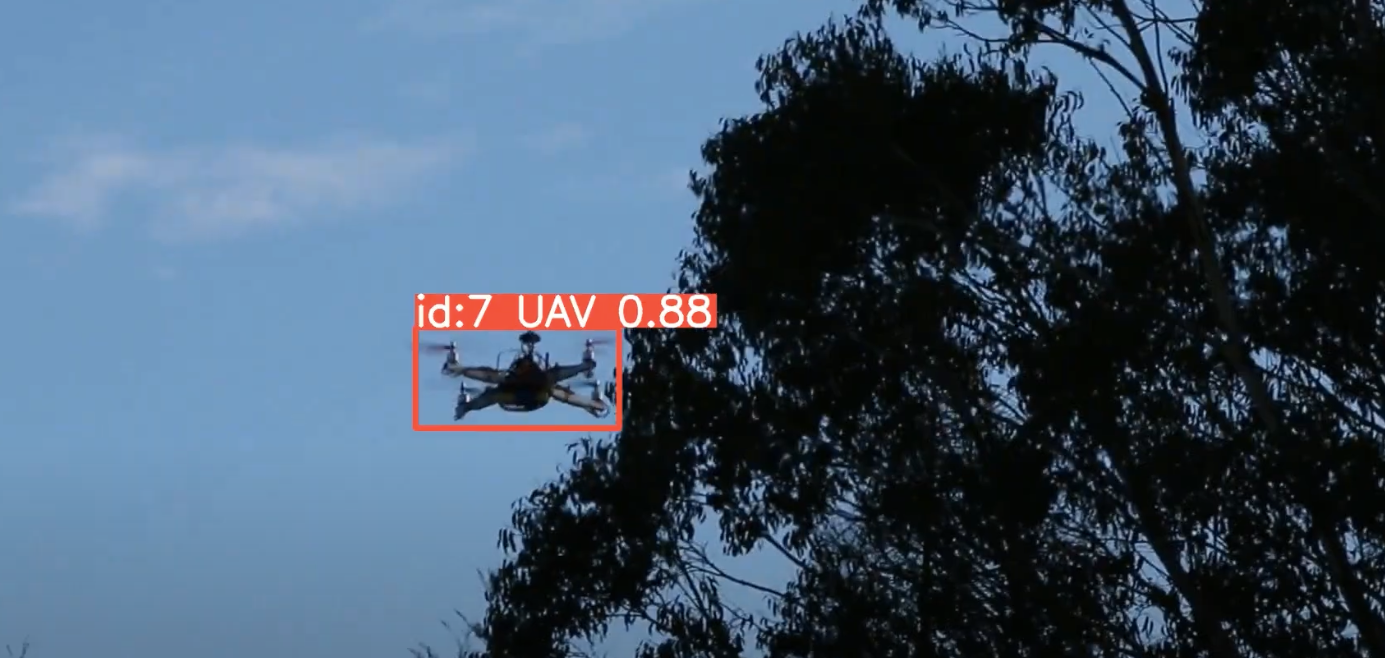}
        \caption{Frame 1}
        \label{fig:yolov5s1-v}
    \end{subfigure}
    \begin{subfigure}{0.2\textwidth}
        \centering
        \includegraphics[scale=0.06]{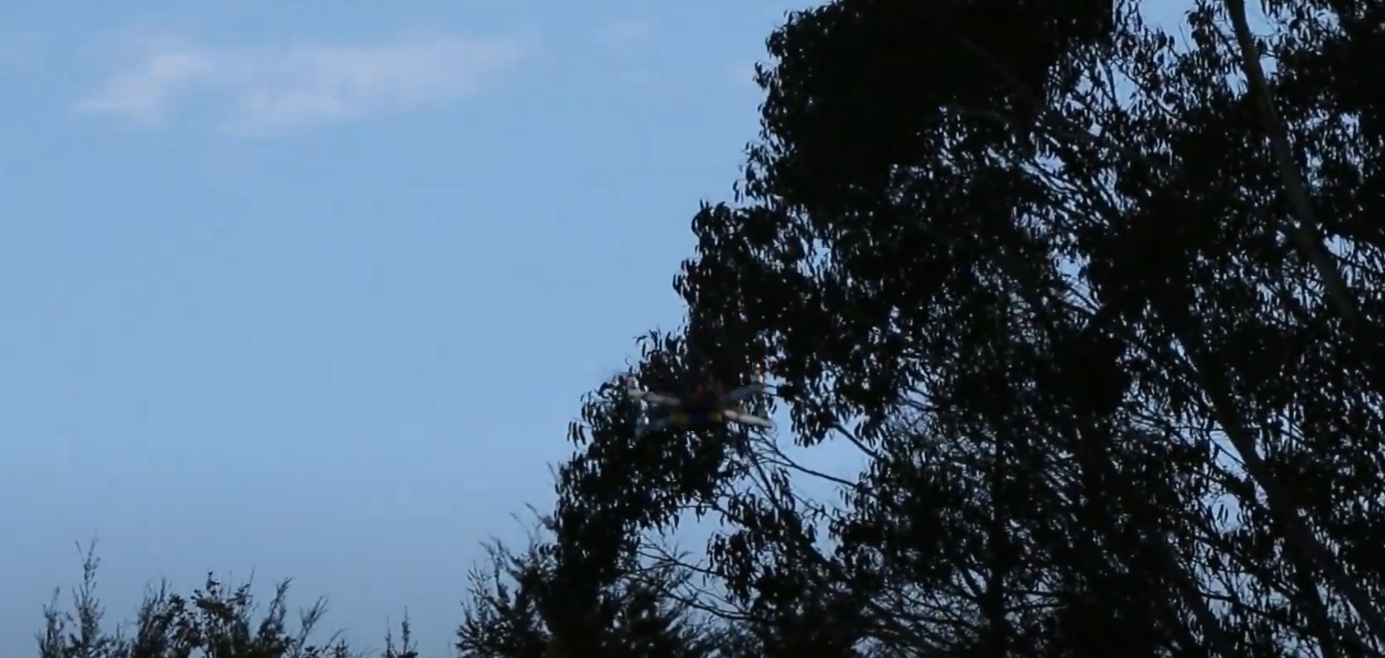}
        \caption{Frame 2}
        \label{fig:yolov5x1-v}
    \end{subfigure}
    \begin{subfigure}{0.2\textwidth}
        \centering
        \includegraphics[scale=0.06]{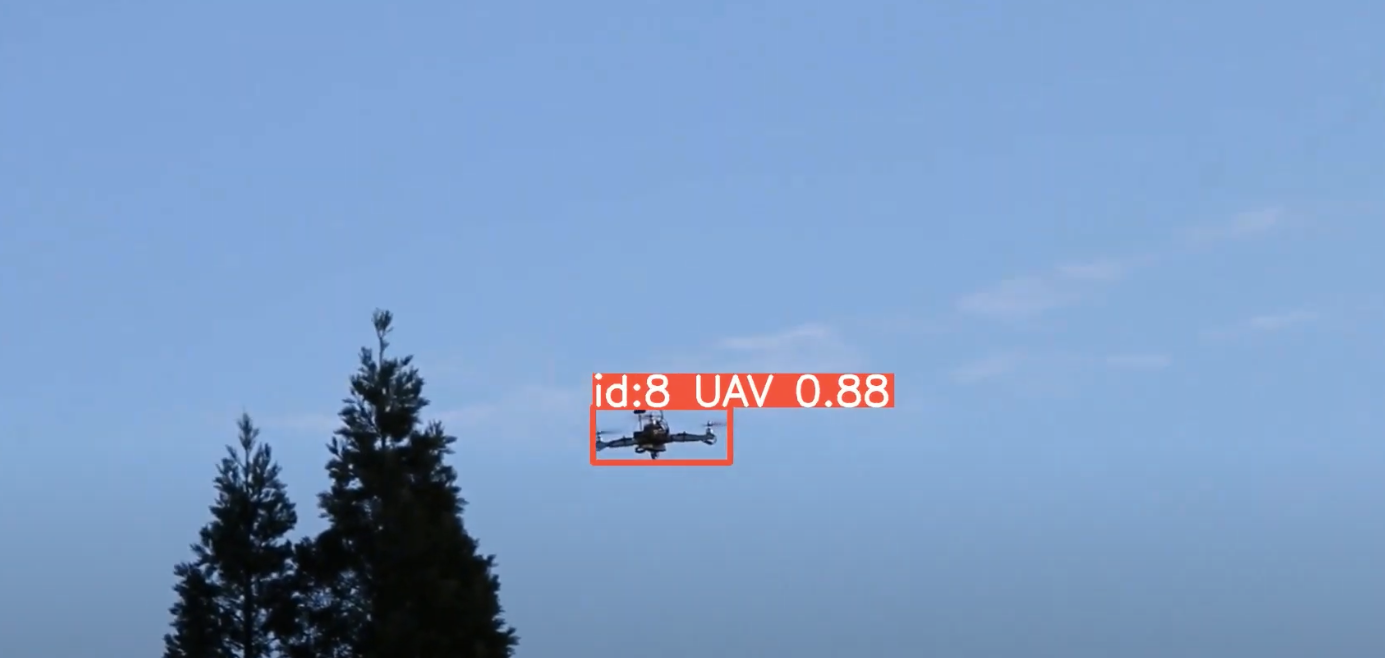}
        \caption{Frame 3}
        \label{fig:yolov8s1-v}
    \end{subfigure}
    \begin{subfigure}{0.2\textwidth}
        \centering
        \includegraphics[scale=0.06]{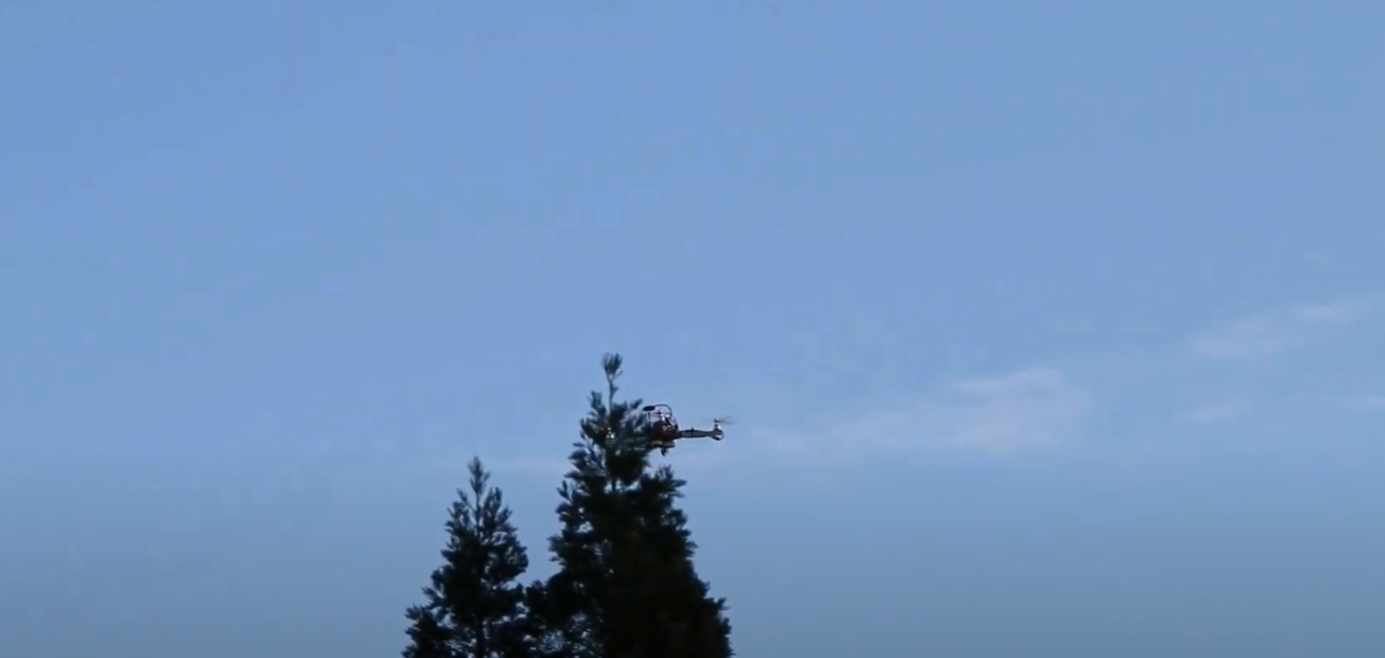}
        \caption{Frame 4}
        \label{fig:yolov8x1-v}
    \end{subfigure}
    \\
    \begin{subfigure}{0.2\textwidth}
        \centering
        \includegraphics[scale=0.06]{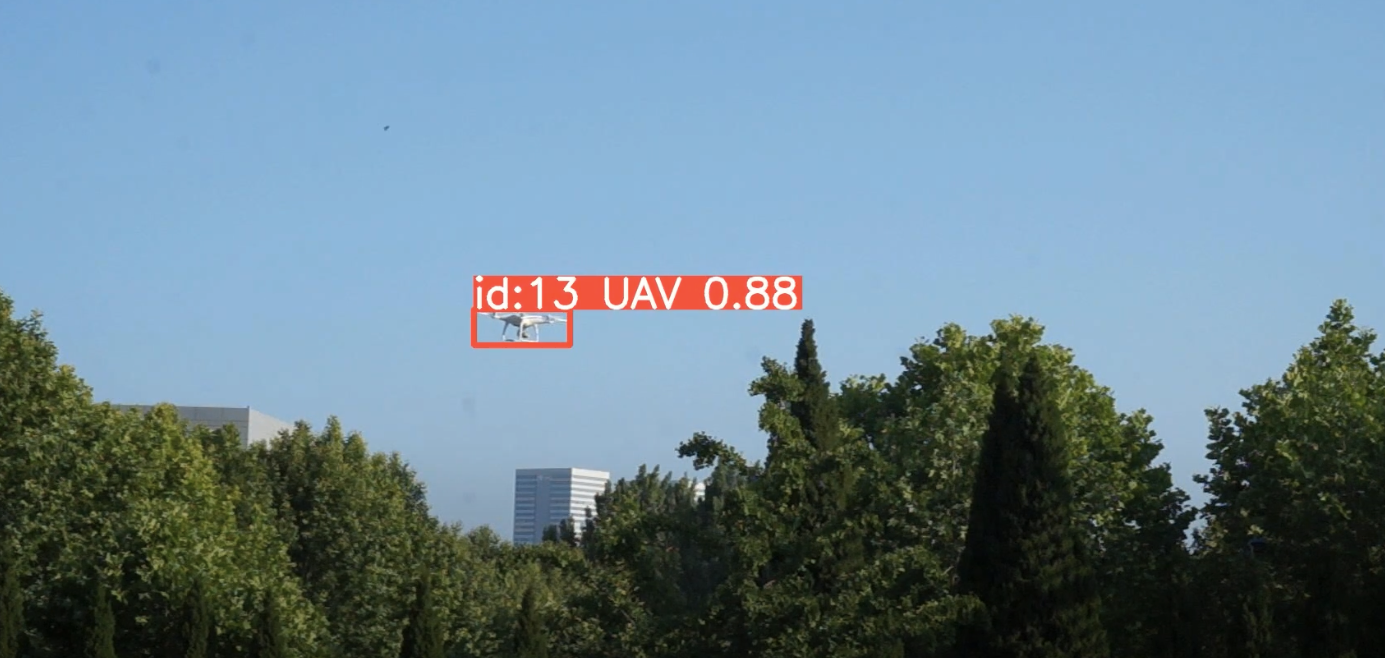}
        \caption{Frame 1}
        \label{fig:yolov5s2-v}
    \end{subfigure}
    \begin{subfigure}{0.2\textwidth}
        \centering
        \includegraphics[scale=0.06]{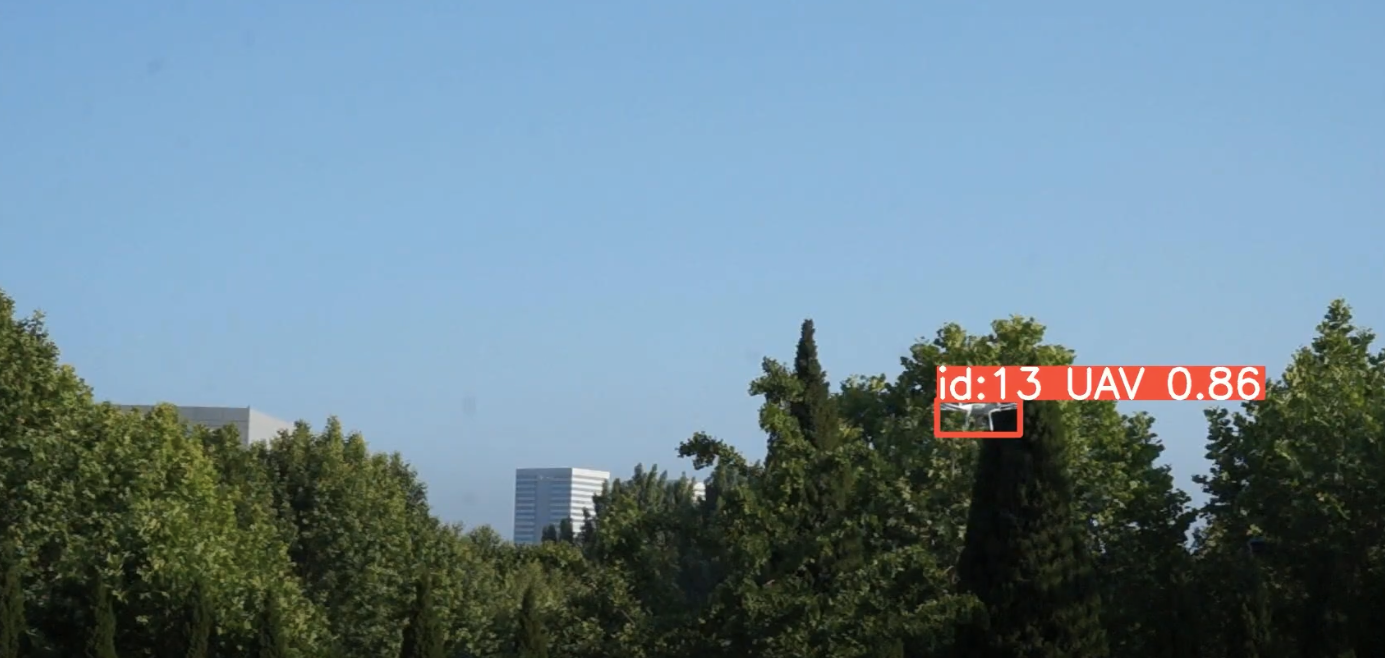}
        \caption{Frame 2}
        \label{fig:yolov5x2-v}
    \end{subfigure}
    \begin{subfigure}{0.2\textwidth}
        \centering
        \includegraphics[scale=0.06]{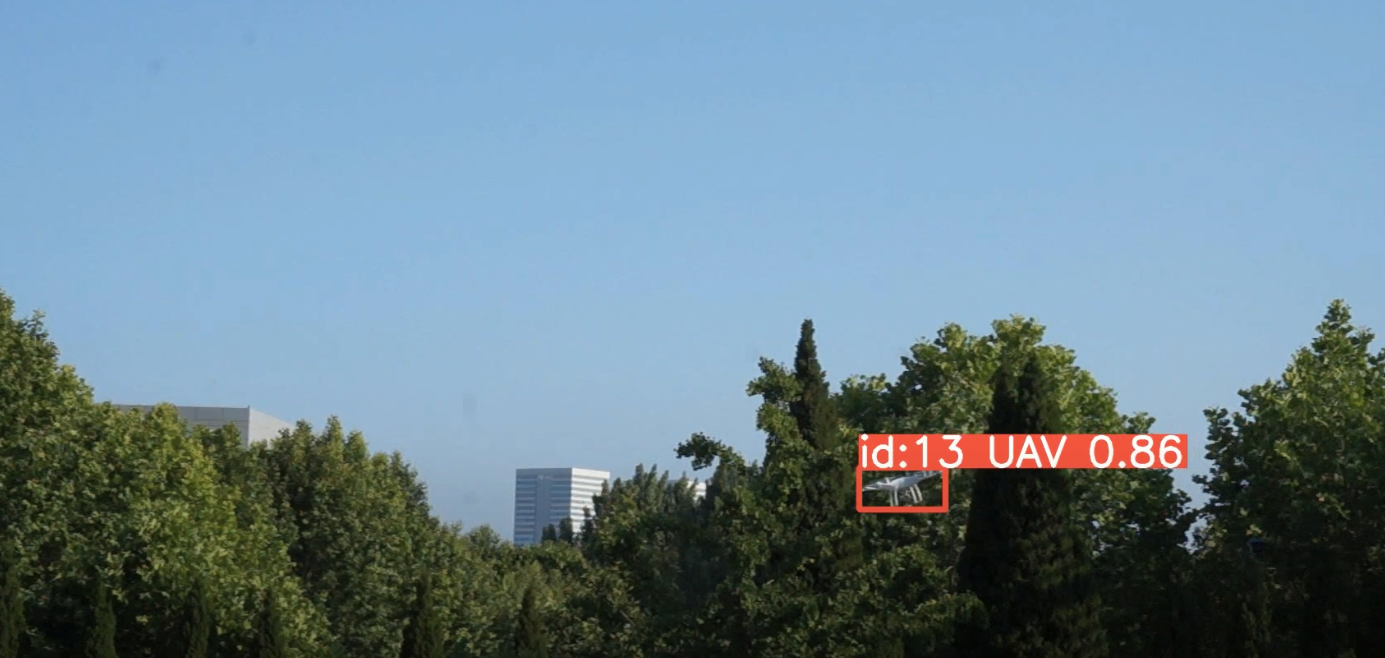}
        \caption{Frame 3}
        \label{fig:yolov8s2-v}
    \end{subfigure}
    \begin{subfigure}{0.2\textwidth}
        \centering
        \includegraphics[scale=0.06]{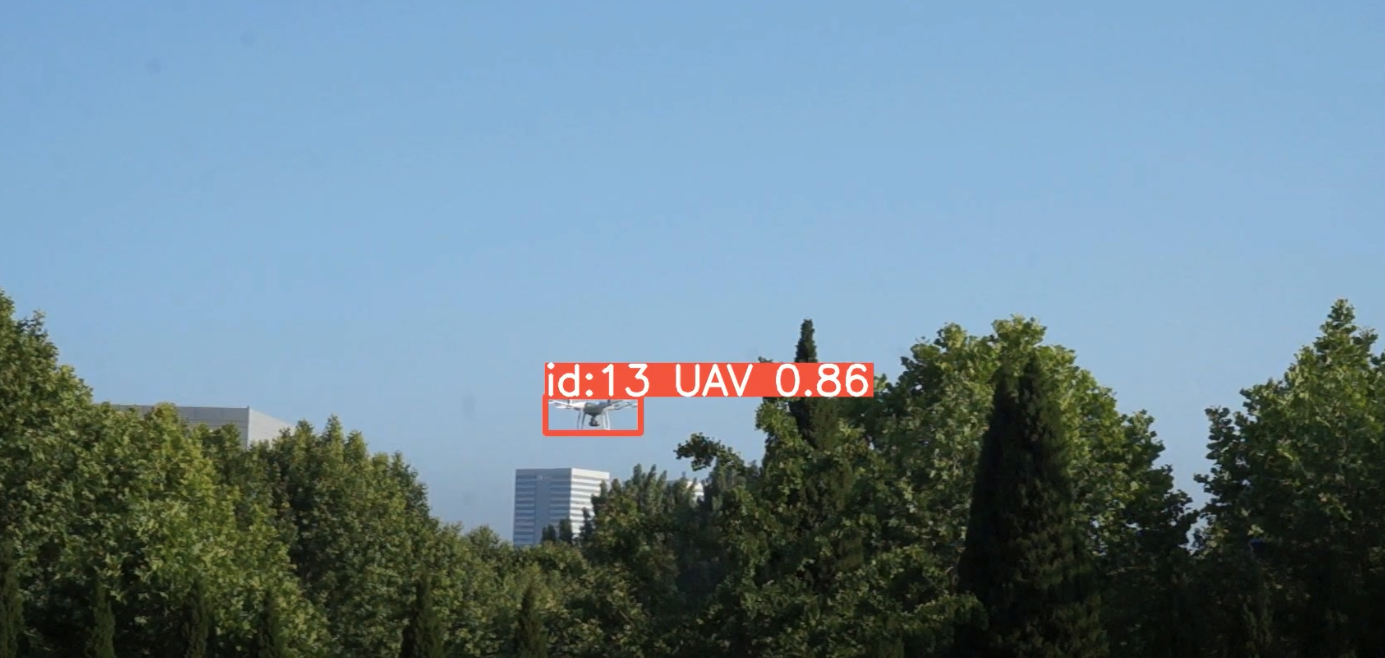}
        \caption{Frame 4}
        \label{fig:yolov8x2-v}
    \end{subfigure}
    \\
    \begin{subfigure}{0.2\textwidth}
        \centering
        \includegraphics[scale=0.06]{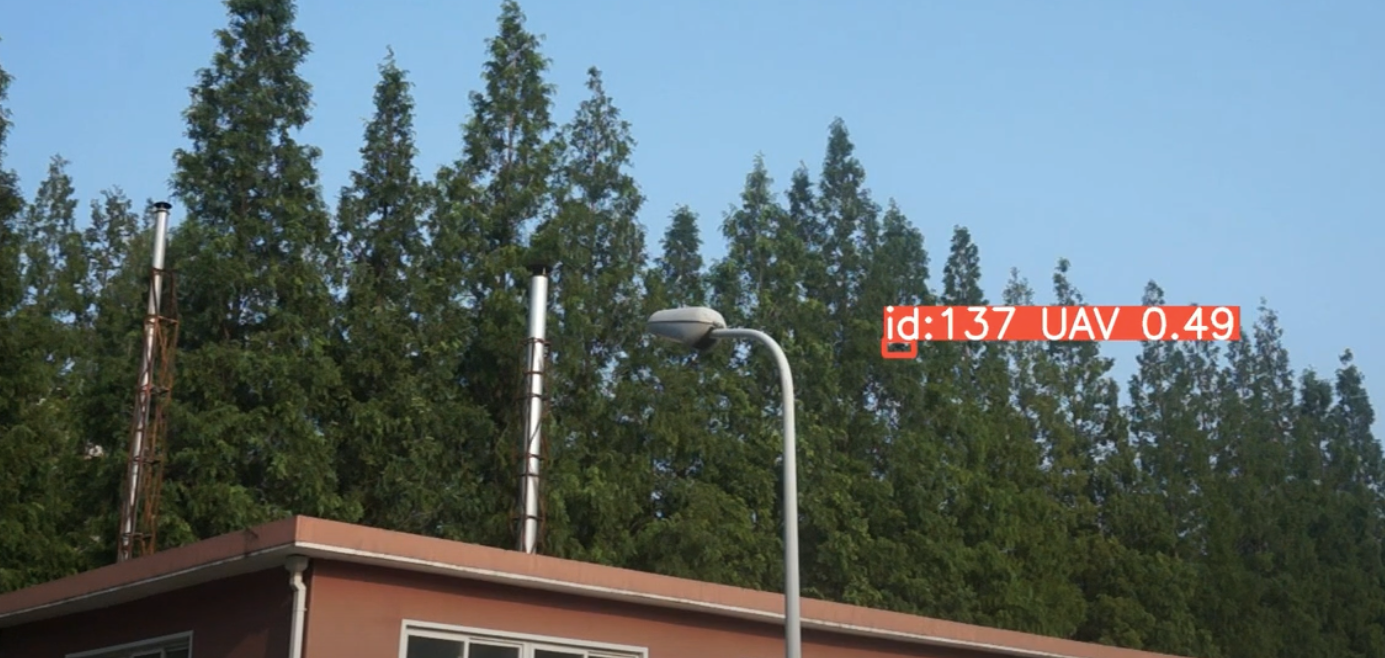}
        \caption{Frame 1}
        \label{fig:yolov5s3-v}
    \end{subfigure}
    \begin{subfigure}{0.2\textwidth}
        \centering
        \includegraphics[scale=0.06]{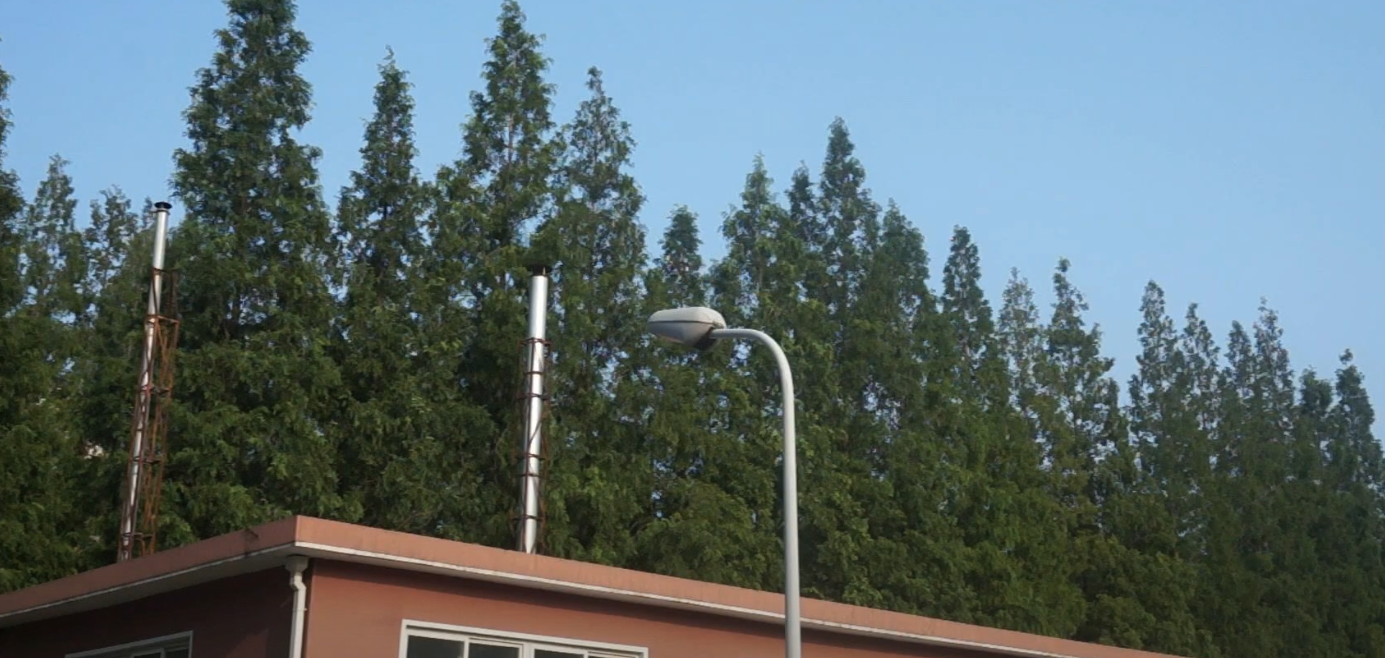}
        \caption{Frame 2}
        \label{fig:yolov5x3-v}
    \end{subfigure}
    \begin{subfigure}{0.2\textwidth}
        \centering
        \includegraphics[scale=0.06]{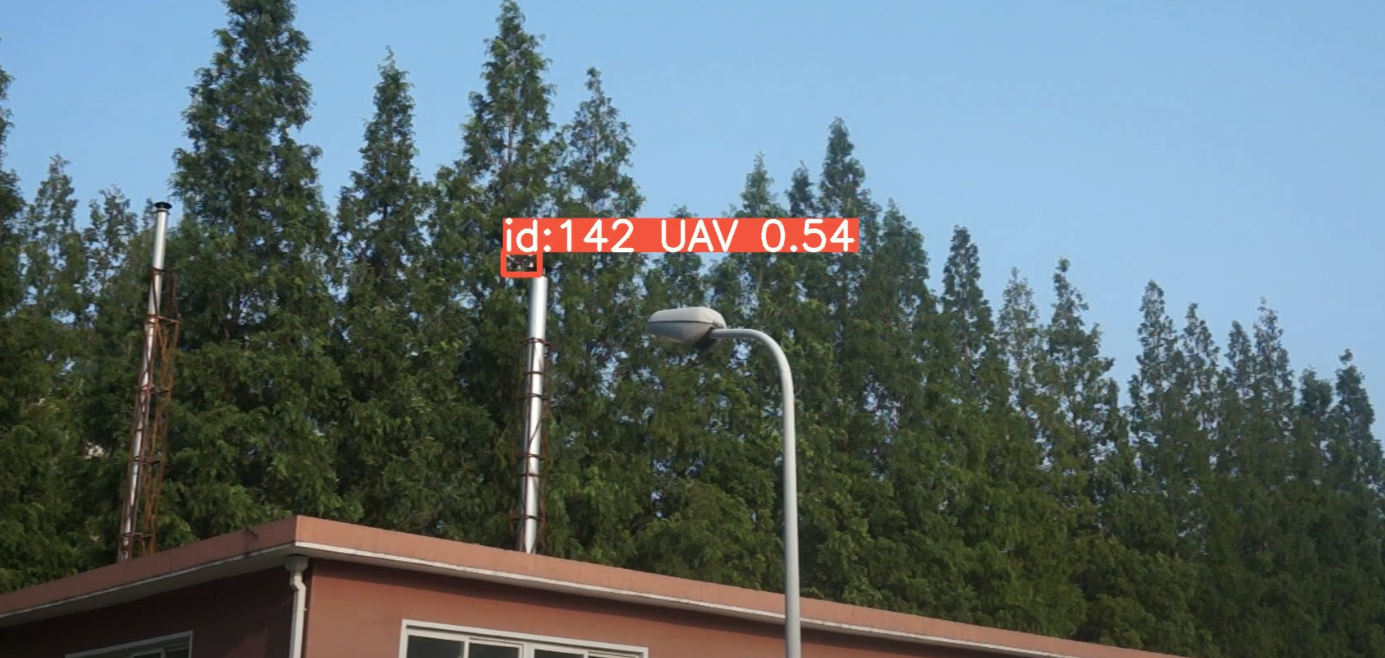}
        \caption{Frame 3}
        \label{fig:yolov8s3-v}
    \end{subfigure}
    \begin{subfigure}{0.2\textwidth}
        \centering
        \includegraphics[scale=0.06]{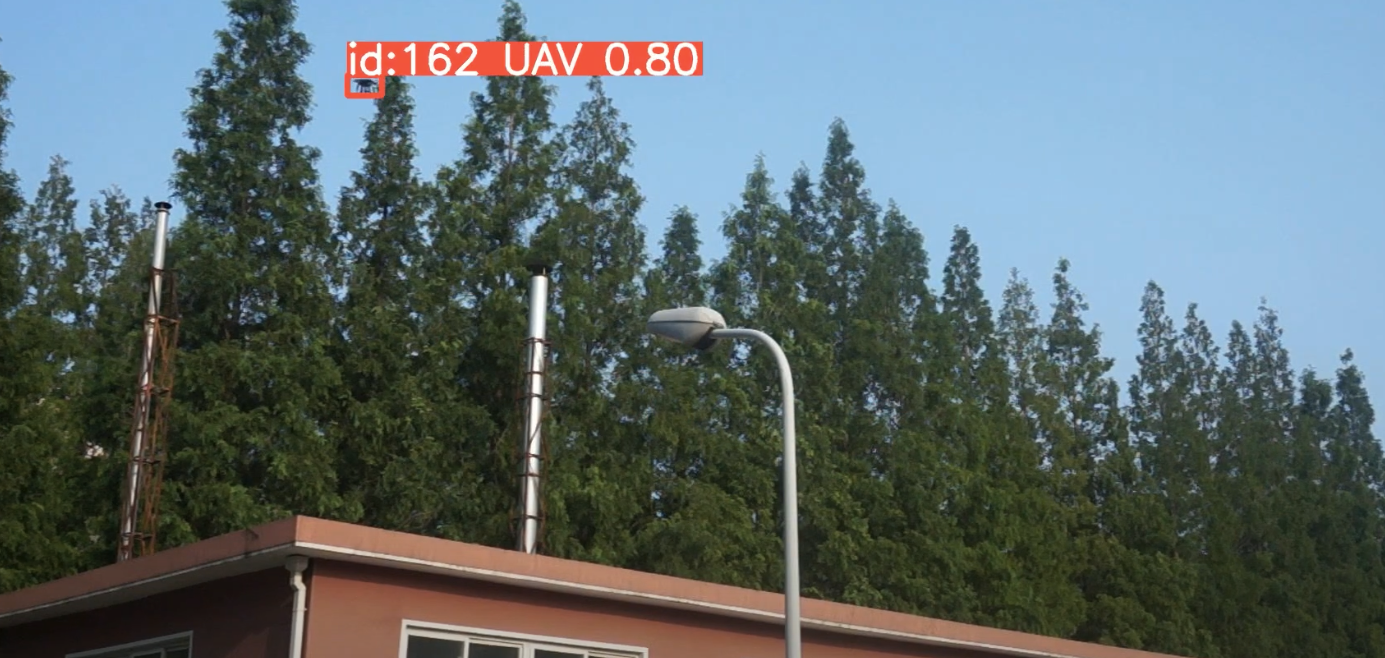}
        \caption{Frame 4}
        \label{fig:yolov8x3-v}
    \end{subfigure}
    \\
    \begin{subfigure}{0.2\textwidth}
        \centering
        \includegraphics[scale=0.06]{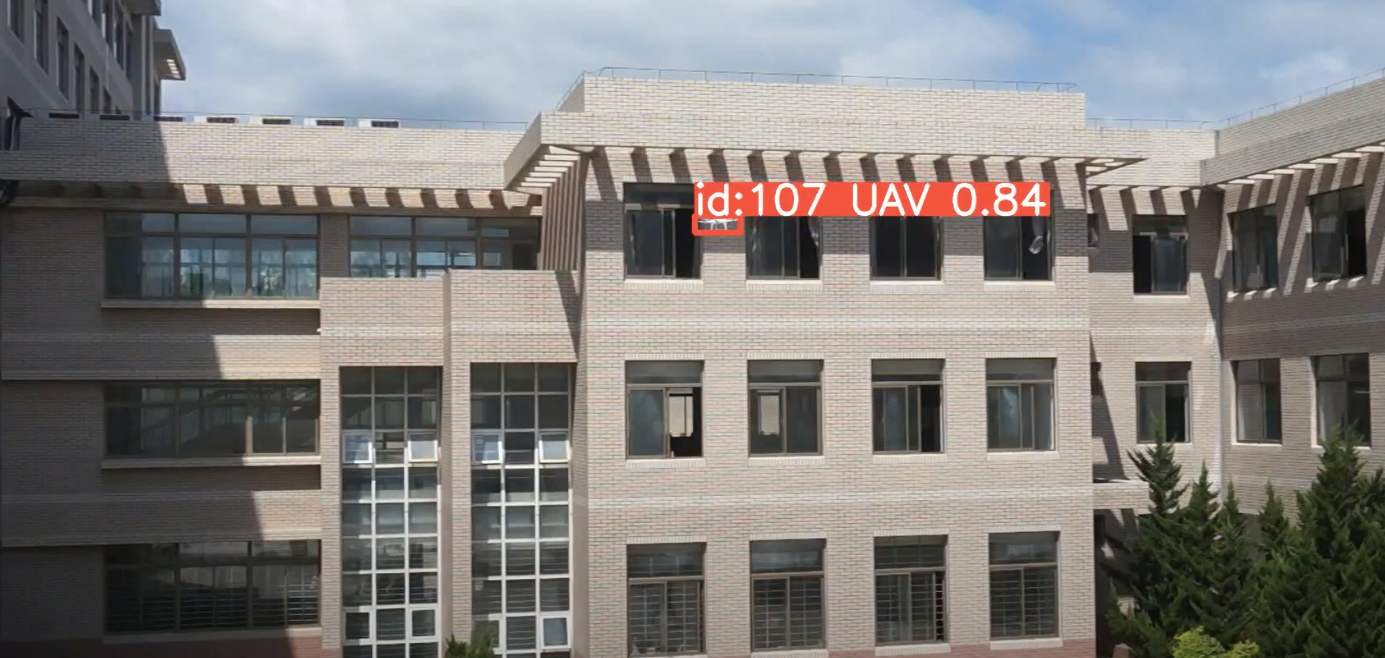}
        \caption{Frame 1}
        \label{fig:yolov5s4-v}
    \end{subfigure}
    \begin{subfigure}{0.2\textwidth}
        \centering
        \includegraphics[scale=0.06]{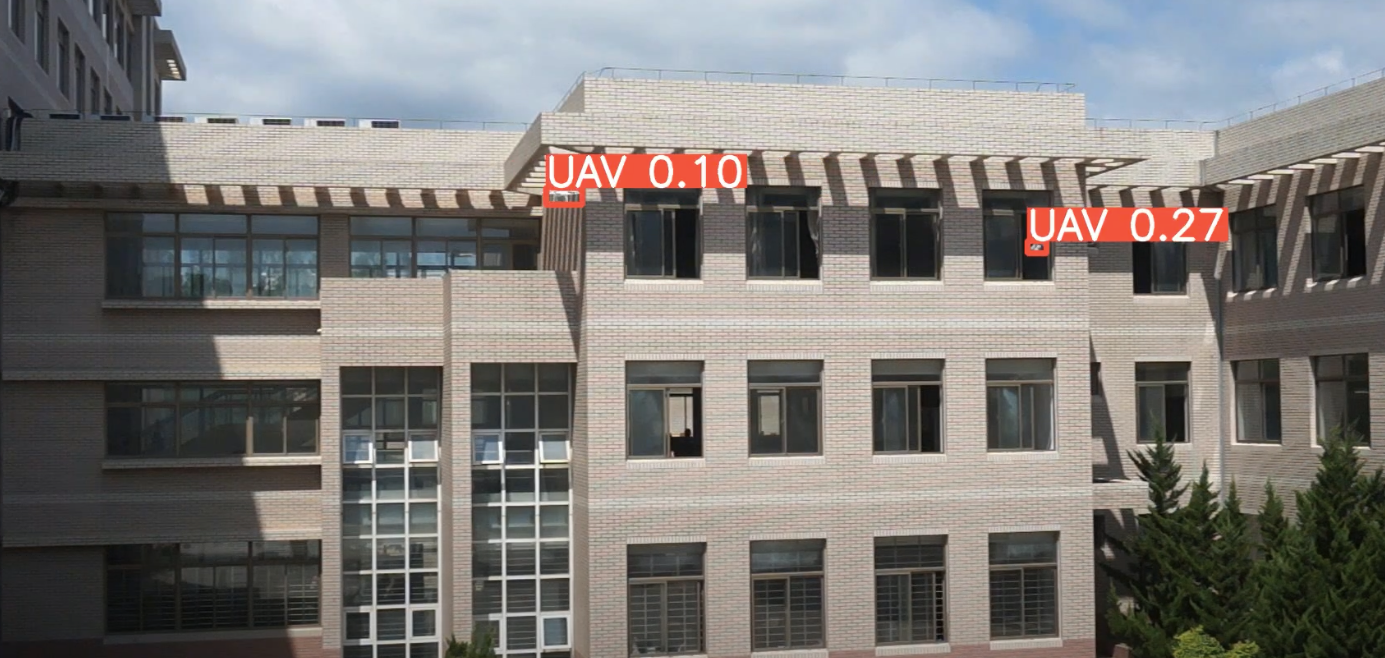}
        \caption{Frame 2}
        \label{fig:yolov5x4-v}
    \end{subfigure}
    \begin{subfigure}{0.2\textwidth}
        \centering
        \includegraphics[scale=0.06]{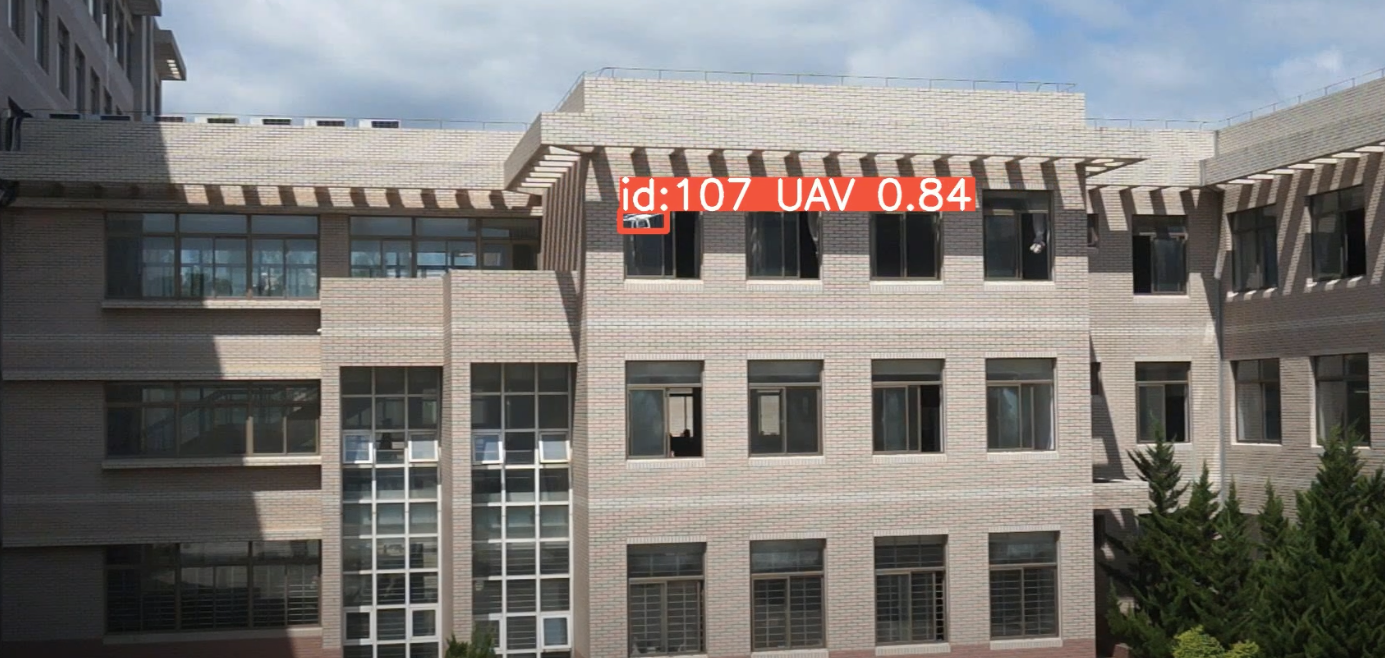}
        \caption{Frame 3}
        \label{fig:yolov8s4-v}
    \end{subfigure}
    \begin{subfigure}{0.2\textwidth}
        \centering
        \includegraphics[scale=0.06]{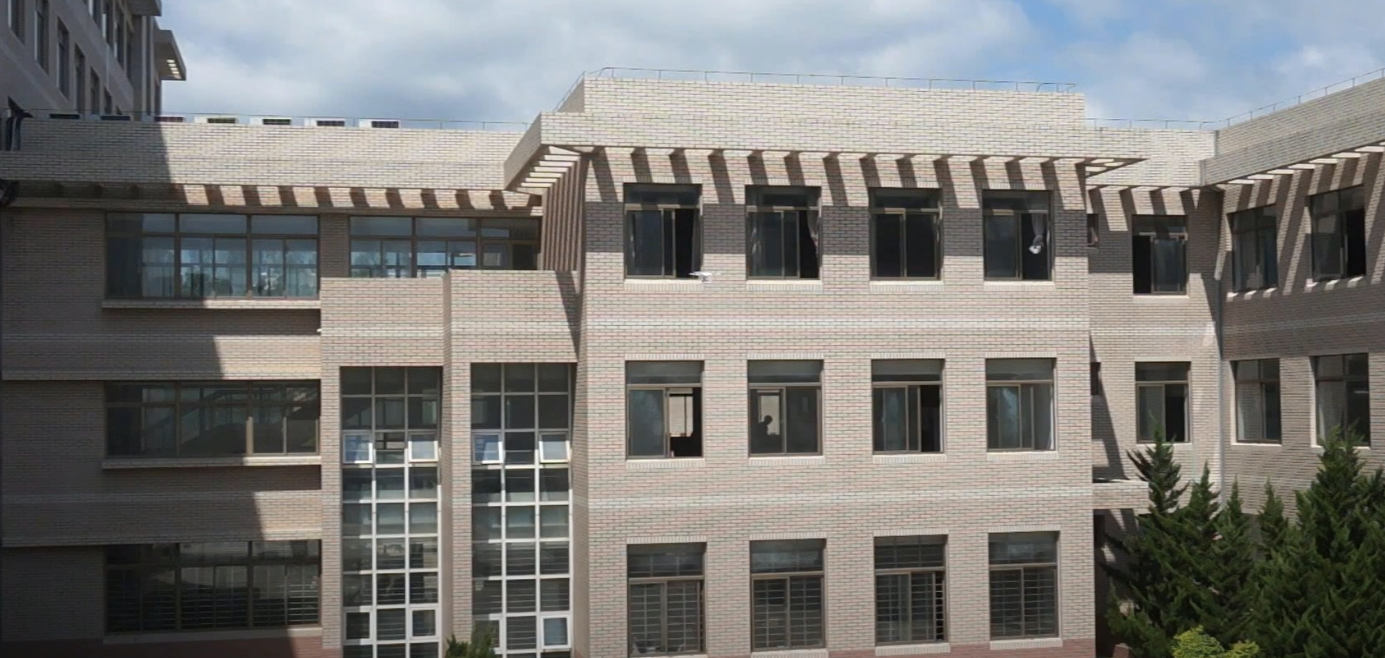}
        \caption{Frame 4}
        \label{fig:yolov8x4-v}
    \end{subfigure}
    \caption{Tracker performance on several testing videos.}
    \label{fig:tracking}
\end{figure}

\subsection{Discussion}
\subsubsection{Data Limitations}
Despite these promising results, we encountered several challenges. One significant limitation is the models' performance in complex environments. While they perform well in controlled conditions, their accuracy drops in cluttered backgrounds (e.g. trees) as seen in Fig. \ref{fig:tracking} (b, d, j, and p) as discussed earlier.

This suggests that further efforts should be done to the datasets. For example, we have seen that most of the models suffered when the quadrotor and the background have comparable contrast values. In other words, if the dataset can have more samples that cover spectrum of contrast values, that will result in better performance of the presented models.

\subsubsection{Precision and Recall vs Confidence Threshold}
Precision measures the percentage of true positive predictions among all positive predictions made by the model, reflecting the accuracy of the positive classifications. Recall, on the other hand, measures the percentage of true positive cases that were correctly identified by the model, indicating the model's ability to capture all relevant instances. The confidence threshold is a parameter that determines the cutoff point at which the model's prediction is considered positive. As clear from figures \ref{fig:yolo_comp} (b and c), increasing the threshold generally leads to higher precision but lower recall, as the model becomes more conservative in making positive predictions. Conversely, lowering the threshold tends to increase recall but reduce precision, as the model includes more positive predictions, some of which may be incorrect. Finding the optimal balance between precision and recall involves selecting an appropriate confidence threshold that aligns with the specific goals and tolerance for error in the application at hand.

\subsubsection{Tracking Metrics}
While Mean IoU and Mean Center Error are essential metrics for evaluating tracking performance, they have certain limitations. Mean IoU may not fully capture the quality of tracking in scenarios with complex object shapes or partial occlusions. Additionally, Mean Center Error, might not reflect the overall accuracy when objects are large or their shapes vary significantly. These metrics do not account for temporal consistency, meaning they do not directly measure how stable the tracking is over time. Therefore, although Botsort demonstrates superior performance in both IoU and center error metrics, further analysis incorporating additional metrics like trajectory smoothness or robustness to occlusions could provide a more comprehensive evaluation of tracking performance.

\subsubsection{Real-Time Deployment}
Real-time processing is another challenge. Although the models demonstrated satisfactory speed, the high computational demands can hinder their deployment in real-time applications. This is particularly true for high-resolution images and video streams, where maintaining a balance between speed and accuracy is critical.

\section{Conclusion}
\label{sec:conclusion}
In this study, the capabilities of advanced deep learning models were explored, specifically YOLOv5 and YOLOv8, for detecting and tracking UAVs. Our comprehensive evaluation on the DUT dataset demonstrated that YOLOv5 models, particularly YOLOv5x, excel in detection accuracy. However, YOLOv8 models showed a remarkable ability to detect less distinct objects, such as blurred images, due to their higher model complexity. Additionally, the performance of two tracking algorithms, Botsort and Byte Track, using metrics such as Mean IoU and Mean Center Error was analyzed. Botsort demonstrated superior performance, achieving higher IoU and lower center error in most cases, indicating more accurate and stable tracking.

In conclusion, while this study has made comparative contribution in UAV detection and tracking techniques, the identified limitations provide valuable insights for future research. By addressing these challenges and exploring the proposed future work directions, we can develop more robust, efficient, and reliable UAV detection systems that enhance safety and security in various applications.





{\small
\bibliographystyle{ieeetr}
\bibliography{ref}
}

\end{document}